\documentclass[letterpaper, 10 pt, conference]{ieeeconf}  %
\usepackage[numbers,sort&compress]{natbib}

\IEEEoverridecommandlockouts                              %

\usepackage{xcolor}
\definecolor{mycitecolor1}{HTML}{668925}
\definecolor{mycitecolor2}{HTML}{E33E33}
\usepackage[pagebackref=false,breaklinks=true,colorlinks,bookmarks=false, citecolor=black, linkcolor=black]{hyperref}
\usepackage{times}
\usepackage{epsfig}
\usepackage{graphicx}
\usepackage{amsmath}
\usepackage{amssymb}

\usepackage{symbols}
\usepackage{tabularx}
\usepackage{multirow}
\usepackage{makecell}
\usepackage{booktabs}

\usepackage{amsmath,amsfonts,bm}

\newcommand{\norm}[1]{\left\lVert#1\right\rVert}

\newcommand{\beas}{\begin{eqnarray*}}
\newcommand{\eeas}{\end{eqnarray*}}
\newcommand{\bea}{\begin{eqnarray}}
\newcommand{\eea}{\end{eqnarray}}
\newcommand{\bes}{\begin{equation*}}
\newcommand{\ees}{\end{equation*}}
\newcommand{\be}{\begin{equation}}
\newcommand{\ee}{\end{equation}}

\def\eg{\emph{e.g}\onedot}
\def\ie{\emph{i.e}\onedot}

\def\1{\bm{1}}

\def\rvc{{\mathbf{c}}}

\def\rvg{{\mathbf{g}}}

\def\rvr{{\mathbf{r}}}

\def\rvx{{\mathbf{x}}}

\def\mA{{\bm{A}}}

\def\mM{{\bm{M}}}

\DeclareMathAlphabet{\mathsfit}{\encodingdefault}{\sfdefault}{m}{sl}
\SetMathAlphabet{\mathsfit}{bold}{\encodingdefault}{\sfdefault}{bx}{n}

\def\gA{{\mathcal{A}}}

\def\gO{{\mathcal{O}}}

\def\gR{{\mathcal{R}}}
\def\gS{{\mathcal{S}}}
\def\gT{{\mathcal{T}}}

\def\gX{{\mathcal{X}}}

\usepackage{color}
\definecolor{Set2c0}{rgb}{0.54117647, 0.16862745, 0.88627451}

\definecolor{gfgg}{RGB}{131, 197, 109}
\definecolor{gfbb}{RGB}{115, 140, 217}
\definecolor{gfrr}{RGB}{208, 119, 111}
\definecolor{gfyy}{RGB}{174, 165, 87}

\definecolor{gggg}{RGB}{188, 228, 208}
\definecolor{bbbb}{RGB}{221, 250, 255}
\usepackage[outline]{contour}

\newcommand*{\ShowNotes}{} %
\definecolor{darkred}{rgb}{0.7,0.1,0.1}
\definecolor{darkgreen}{rgb}{0.1,0.7,0.1}
\definecolor{cyan}{rgb}{0.7,0.0,0.7}
\definecolor{dblue}{rgb}{0.2,0.2,0.8}
\definecolor{maroon}{rgb}{0.76,.13,.28}
\definecolor{burntorange}{rgb}{0.81,.33,0}
\definecolor{tealblue}{rgb}{0.212,0.459, 0.533}

\definecolor{pp}{rgb}{0.43921569, 0.18823529, 0.62745098}
\definecolor{rr}{rgb}{0.5254902 , 0.00784314, 0.12941176}
\definecolor{bb}{rgb}{0.09019608, 0.23529412, 0.37647059}
\definecolor{yy}{rgb}{0.49803922, 0.3372549 , 0.0}
\definecolor{gg}{rgb}{0.02352941, 0.3372549 , 0.17647059}

\ifdefined\ShowNotes
  \newcommand{\colornote}[3]{{\color{#1}\bf{#2: #3}\normalfont}}
\else
  \newcommand{\colornote}[3]{}
\fi

\newcommand{\eat}[1]{} %

\title{\LARGE \bf
Semantic Tracklets: An Object-Centric Representation for \\ Visual Multi-Agent Reinforcement Learning
}

\author{Iou-Jen Liu$^{*1}$,  Zhongzheng Ren$^{*1}$,  Raymond A. Yeh$^{*1}$,  Alexander G. Schwing$^{1}$
\thanks{$^{1}$University of Illinois at Urbana-Champaign, IL, USA.}
\thanks{\tt\small iliu3@illinois.edu}
\thanks{$^*$ Equal contribution.}
}

\renewcommand{\citep}[1]{\cite{#1}}
\renewcommand{\citet}[1]{\cite{#1}}
\begin{document}

\maketitle
\thispagestyle{empty}
\pagestyle{empty}

\begin{abstract}
Solving complex real-world tasks, \eg, autonomous fleet control, 
often involves a coordinated team of multiple agents which learn strategies from visual inputs via reinforcement learning. Many existing multi-agent reinforcement learning (MARL) algorithms however don't scale to environments where agents operate on visual inputs. 
To address this issue, algorithmically, recent works have focused on  non-stationarity and exploration. 
In contrast,  we study whether  
scalability can also be achieved via a disentangled representation. %
For this, we  explicitly construct  an object-centric intermediate representation to characterize the states of an environment, which we refer to as 
`semantic tracklets.' We evaluate `semantic tracklets' on the visual multi-agent particle environment (VMPE) %
and on the challenging visual multi-agent GFootball environment. %
{`Semantic tracklets' consistently outperform baselines on VMPE, and achieve
a {$+2.4$ higher score difference than baselines} on GFootball.} Notably, this method is the first to successfully learn a strategy for five players  in the GFootball environment using only visual data. For more, please see our project page: \url{https://ioujenliu.github.io/SemanticTracklets}
\end{abstract}

\section{Introduction}
\label{sec:Intro}%
Many real-world tasks, such as autonomous fleet control~\cite{Mariani20} and swarm robot control~\cite{Mohan09, Schranz20}, are naturally modeled as  visual multi-agent systems. In these systems, multiple agents learn to coordinate based on pixel inputs.

Many existing works in multi-agent reinforcement learning (MARL) study learning of multi-agent coordination~\cite{maddpg, Bargiacchi18, Raileanu18, maven, pic, qmix, Yang18, coma, Kim19, Das19}.  In common to all these works is the use of a compact observation vector which summarizes the situation for each agent. For instance, on
the simulated particle world~\cite{Mordatch17} the  Multi-Agent Actor-Critic~\cite{maddpg} method  uses the locations and velocities of controlled  and neighboring particles as input to achieve compelling results, albeit, training times are often significantly longer than those of single-agent reinforcement learning. 

Despite this success, few works for MARL focus on visual agents which act purely upon image observations. 
Indeed, Visual Multi-Agent Reinforcement Learning (VMARL) which operates on pixel inputs %
adds additional complexity to the already challenging MARL task, increasing training times often significantly. Unsurprisingly, current VMARL methods trained end-to-end with conv-nets hardly learn useful polices in complex environments. %
For example, in the Google Research Football (GFootball) environment, end-to-end training with conv-nets doesn't result in meaningful policies and %
remains an open problem~\cite{gfootball}. %
To tackle this challenge, existing works primarily use imitation learning to reduce the long training times. For example, Jain~\etal~\citet{Jain19}, who study a collaborative navigation task {involving up to three agents}, first train with imitation learning~\cite{dagger} followed by RL fine-tuning. %
However, imitation learning requires ``expert'' demonstrations that may be hard to obtain. This is particularly true  in complex MARL environments like GFootball, where strategies are necessary. %
{Note, a ground-truth ``expert'' strategy may not even exist.} 

\begin{figure}[t]
\centering
\includegraphics[width=0.5\textwidth]{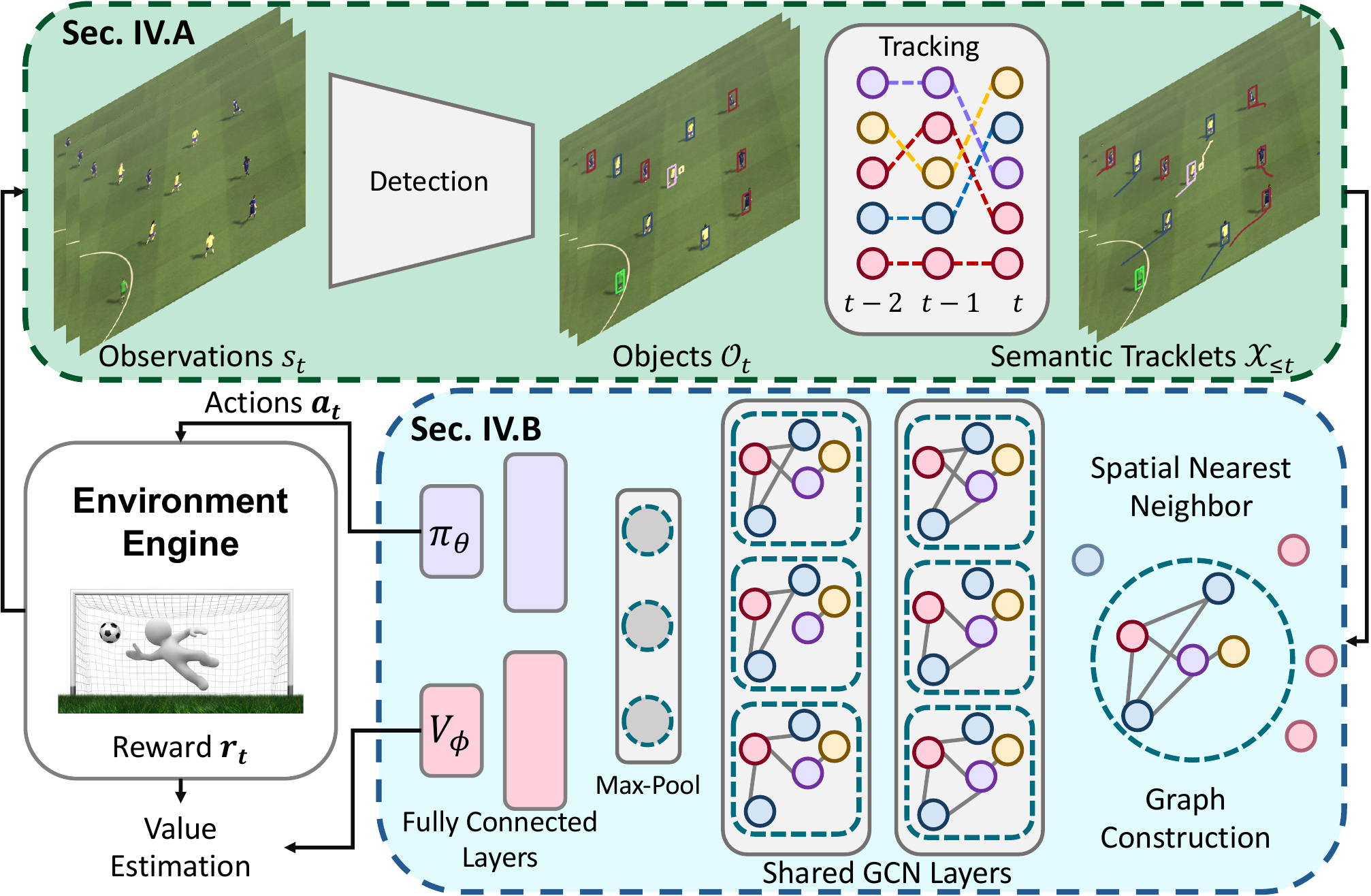}
\caption{The proposed visual MARL system illustrated using the GFootball environment as an example. Given image observations, we first perform object detection to identify the objects (\eg, ball,  players on the home  and the visiting team shown in {\bf \color{yy}yellow}, {\bf \color{rr}red} and {\bf \color{bb}blue}). We perform tracking to maintain the identity of the detected objects across frames before extracting  semantic tracklets. Next, we construct a graph representation using the $K$-nearest neighbors around the  controlled agent (active player), shown in {\bf \color{pp}purple}. Lastly, we compute the policy and value function via a graph convolutional  net (GCN) and interact with the  environment. %
}
\vspace{-0.5cm}
\label{fig:pipeline}
\end{figure}

In this paper, we study an alternative approach to make VMARL more scalable.  We design an intermediate representation for more efficient training by building  helpful inductive biases into the models. For this we propose `semantic tracklets,' an object-centric representation to characterize the surroundings of each agent. Different from imitation learning, this approach does not require ``expert'' demonstrations, \ie, \emph{annotations in the form of actions}. Instead, it uses the more accessible alternative of  \emph{annotations in the form of object labels}, which can be collected by ``anyone'' without domain knowledge of the underlying task.

At a high-level and pictorially sketched in \figref{fig:pipeline}, semantic tracklets contain two components: the `semantics' which answer ``what'' is the object; and the `tracklets'  which answer ``where'' is the object. This representation provides a useful inductive bias for VMARL, where the agents correspond to an ``object'' in the given pixel inputs, and agents'  interaction with the environment involves spatial movements.  
Furthermore, such an object-centric representation enables the use of graph-nets which naturally model coordination/interactions among the agents while being invariant to the agents' ordering, an important property for homogeneous agents~\cite{pic, Jiang19}. %

We evaluate our approach on two environments, the Visual Multi-Agent Particle  Environment (VMPE), and the challenging Google Research Football Environment (GFootball) with only visual input~\cite{gfootball}. 
On VMPE, we compare semantic tracklets with a baseline  that directly uses pixel inputs, and a baseline that uses semantic segmentation as an intermediate representation~\cite{Hong17}.  %
{Semantic tracklets consistently outperform baselines on VMPE.} 
On GFootball {\em 11 \vs 11} full game, semantic tracklets achieve a $+2.4$ higher score difference than baselines. %
We show that `semantic tracklets' permit to  learn, for the first time, collaboration among  five players in the GFootball environment using only visual input. The players learn to cooperate and outperform their opponent.

\section{Related Work}
\label{sec:rel}

{\noindent \bf Intermediate Representations.}
In reinforcement learning, the representation which characterizes the state of an environment has a direct impact on the accuracy and efficiency of the learned policy. While end-to-end learning-based methods, which learn a policy from the pixel space~\cite{dqn1, dqn2, a3c, ppo, kf, iswitch}, have achieved impressive results, others have demonstrated that designing intermediate representations may be beneficial~\cite{Hong17, Mousavian2019Object}. For example, Hong~\etal ~\cite{Hong17} study the use of semantic segmentation to tackle the `virtual-to-real'~\cite{Tan18, James19, Hong17} problem in single-agent robot navigation tasks. 

More recently, object-centric representations demonstrate promising results in various domains including robot learning~\cite{devin2018deep} and autonomous-driving~\cite{ye2020object}.
For example, Ye~\etal~\cite{ye2020object} build an object-centric representation for learning polices to perform robotic manipulation tasks.
Generally, these approaches rely on object detection, trained with annotated data, %
to maintain a structured representation of the objects.

We note that the aforementioned methods consider only single-agent settings. 
The interactions between multiple controlled agents, which are critical for multi-agent learning, are not discussed.
Moreover, some of the existing works~\cite{ye2020object} %
rely on supervised learning or imitation learning to train policies. 
Different from these works,  we study an object-centric representation for VMARL. %
The proposed semantic tracklets permit  to capture the interaction of multiple agents via graph-nets and to learn cooperative policies efficiently.

\noindent \textbf{Multi-Agent RL.} To efficiently learn policies in multi-agent systems, a variety of multi-agent RL algorithms have been proposed~\cite{maddpg, Bargiacchi18, Raileanu18, maven, pic, qmix, Yang18, coma, Kim19, Das19, hts-rl, Jain21, cmae, CordialSync}. For example, to cope with  non-stationarity, `Multi-agent Actor-critic'~\cite{maddpg} uses a centralized critic which operates on all agents' observations and actions. %
`Monotonic Value Function Factorisation'~\cite{qmix}  advocates estimating joint action-values as a non-linear combination of per-agent values.
However, all  approaches assume a compact feature vector as input. Scaling of MARL approaches to agents operating with visual observations remains open in those works. 

\noindent \textbf{Visual Multi-Agent RL.} 
Jain \etal~\citet{Jain19} study interaction of two visual agents in AI2Thor~\cite{ai2thor}. For training to scale, imitation learning~\cite{dagger} is used initially, which requires access to an `expert,' \ie, \emph{annotations for actions}. For complex environments and particularly for multi-agent tasks, such an `expert' may not be available. Hence, we study an alternative: building an object-centric representation to learn the policies efficiently. %
For this our method only requires  \emph{annotations of  objects}, which are obtained more easily or are even already available from off-the-shelf detectors.

Visual-rich environments, \eg, AI2Thor~\cite{ai2thor}, Habitat~\cite{habitat}, and iGibson~\cite{igibson}, permit to study embodied AI tasks, such as navigation and question answering.  While some of the environments~\cite{ai2thor, igibson} have recently been extended to support multi-agent training, the involved tasks are often light on collaboration. This is largely due to the fact that navigation is considered important, which leaves little room for collaboration. 
For this reason, here, we consider GFootball~\cite{gfootball}, where agents are required  to play soccer  
given as observation rendered game frames. 
The environment supports both multi-agent and single agent settings of visual RL.  
The model can control any number of players on a team,  from one to all players, making it particularly suitable to study  collaboration.

For GFootball, Kurach~\etal~\citet{gfootball} consider controlling only one visual agent in the {\em 11 \vs 11} full game. %
The method of Kurach~\etal~\cite{gfootball} requires 500 parallel actors and more than 100 million environment training steps. In contrast, with `semantic tracklets,' we successfully learn polices to control up to five visual agents in the {\em 11 \vs 11} full game %
within 20 million environment steps. %

\section{Preliminaries}
\label{sec:prelim}
RL studies how agents should interact with an environment to 
maximize their expected future rewards. %
We first provide background on single- and multi-agent RL.

\noindent\textbf{Single-Agent RL.} 
An environment is commonly modeled as a Markov Decision Process (MDP). Formally, an MDP is defined by:  the environment's state space $\gS$,   the set of actions $\gA$ which an agent can perform,   a transition function $\gT: \gS \times \gA \mapsto \Delta_\gS$ %
specifying for each state the probability with which it   is reached next when performing an action in the current state, and  a reward function $\gR: \gS \times \gA \mapsto \mathbb{R}$ denoting the prize  for executing an action in a given state. 

An agent's interaction with the environment is described via a  policy $\pi_\theta$, parameterized by $\theta$. Given the  state $s_t \in \gS$ at time step $t$,  the policy 
models the probability of performing an action $a$, \ie, $\pi_\theta (a|s_t) \in [0,1]$ and $\sum_{a\in\gA} \pi_\theta (a|s_t) =1$.

RL aims to find the  policy $\pi_\theta$ that maximizes the expected return
$J(\theta) \triangleq \mathbb{E}_{s_t\sim\rho^{\pi_\theta}, a_t\sim \pi_\theta} \left[{R}_t \right]$, %
where ${R}_t \triangleq \sum_{k=0}^K \gamma^{k} r_{t+k}$ is the expected discounted return. %
Here, $K \in \mathbb{Z}^+$ denotes the time horizon, $\rho^{\pi_\theta}$ is the state distribution induced by $\pi_\theta$, $r_{t+k} = \gR(s_{t+k}, a_{t+k})$, and $\gamma$ is a discount factor. 

\noindent\textbf{Proximal Policy Optimization (PPO).}
To learn the parameters $\theta$ of the policy $\pi_\theta$, PPO~\cite{ppo} is a commonly used algorithm. PPO
employs the  `surrogate' objective function
\bea
\label{eq:ppo_h}\nonumber
  \hat{J}_{\text{PPO}}(\theta) =
   \mathbb{E}_{(s_t, a_t, r_t) \sim {\rho_{\pi_\theta}}}[
   \min(\xi_t(\theta)D_t,\\ \bm{[\![}\xi_t(\theta)\bm{]\!]}_{1-\epsilon}^{1+\epsilon} D_t)
    + \lambda H(\pi_{\theta}(\cdot|s_t))],
\eea
where $\bm{[\![ \cdot }\bm{]\!]}_a^b$ denotes clipping to interval $[a,b]$, and $\xi_t(\theta) = \frac{\pi_{\theta}(a_t|s_t)}{\pi_{\theta_\text{old}}(a_t|s_t)}$ denotes the probability ratio of the current policy $\pi_{\theta}(a_t|s_t)$ to a slightly outdated policy $\pi_{\theta_\text{old}}(a_t|s_t)$. %
$D_t = R^{(n)}_t  - V_{\phi}(s_t)$ is the estimated advantage function, where $R^{(n)}_t = \sum_{k=0}^{n-1} \gamma^kr_{t+k} + \gamma^nV_{\phi}(s_{t+k})$ is the $n$-step  truncated return, and $V_{\phi}$ is the value function parameterized by $\phi$. 

Intuitively, this surrogate objective function encourages to learn a good policy, while
preventing policy collapse by ensuring that the new policy does not deviate too much from the old policy. Additionally, the entropy term $H$ is added to encourage exploration. The parameters $\phi$ of the value function $V_{\phi}$ are learned by minimizing the squared loss $J_V(\phi) = \mathbb{E}_{(s_t, r_t) \sim {\rho_{\pi_\theta}}}[(R_t^{(n)} - V_{\phi}(s_t))^2] $. 

\noindent\textbf{Multi-agent RL.}
A multi-agent MDP with $N$ agents consists of 
a state space $\gS$, a transition function $\gT$, a set of reward functions $\{\gR^i\}_{i=1}^N$, 
and a set of action spaces $\{\gA^i\}_{i=1}^N$, where ${\gR}^i$ and ${\gA}^i$ correspond to reward and action space of agent $i$. The transition function ${\gT}: {\gS} \times {\gA}^1 \dots \times {\gA}^N \mapsto {\Delta_\gS}$ maps the current state and actions taken by all $N$ agents to a next state.  

At each time step $t$, each agent $i$ takes action $a^i_t$ and receives state $s_{t+1}$ and reward $r^i_t$ from $\gR^i: \gS \times \gA^1 \times \dots \times \gA^N \mapsto \mathbb{R}$. We denote all actions and rewards at time $t$ by $\bm{a_t} = (a^1_t, \dots, a^N_t)$ and $\bm{r_t} = (r^1_t, \dots, r^N_t)$. Note, most MARL works~\cite{qmix, maddpg, coma} assume each agent receives an individual compact local observation. In contrast, in our setting, each agent only has access to the same rendered image, denoted by $s_t$. 
{Operating on the same observation is often a more  challenging setting than having access to individual local observations. For instance, in many video games, \eg, Starcraft 2, Dota, and Age of Empires, a local first-person visual observation for individual units is not available. In those games, players only have access to the game image. Moreover, 
in a fulfillment center, low-cost robots can be controlled using classical security camera footage.
}

Consider $N$ policies $\Pi=\{\pi_{\theta^{i}}\}_{i=1}^N$ with parameters $\{\theta^i\}_{i=1}^N$, 
and $N$ value functions $\{V_{\phi^{i}}\}_{i=1}^N$ with parameters $\{\phi^i\}_{i=1}^N$, which are associated with the $N$ policies. Multi-Agent Actor-Critic~\cite{maddpg} extends the objective function of PPO, given in \equref{eq:ppo_h}, to read  
\bea
\label{eq:mappo}\nonumber
  \hat{J}_{\text{MAPPO}}(\theta^i) =
   \mathbb{E}_{(s_t, \bm{a_t}, \bm{r_t}) \sim {\rho_{{\Pi}}}}[\min(\xi^i_tD^i_t,\\
   \bm{[\![}\xi_t^i(\theta)\bm{]\!]}_{1-\epsilon}^{1+\epsilon}D^i_t)
   + \lambda H(\pi_{\theta^i}(\cdot|s_t))]. 
\eea
Here $\rho_{{\Pi}}$ denotes the trajectory distribution induced by ${\Pi}$, and $\xi^i_t = \frac{\pi_{\theta^i}(a^i_t|s_t)}{\pi_{\theta^i_\text{old}}(a^i_t|s_t)}$ is the probability ratio.
$D^i_t = R^{i,(n)}_t  - V_{\phi^{i}}(s_t)$ %
is the estimated advantage function, where $R^{i,(n)}_t = \sum_{k=0}^{n-1} \gamma^kr^i_{t+k} + \gamma^nV_{\phi^{i}}(s_{t+k})$.  %
Value function $V_{\phi^{i}}$ is updated by minimizing $J_V(\phi^i) = \mathbb{E}_{(s_t, r^i_t) \sim {\rho_{\bm{\Pi}}}}[(R_t^{i,(n)} - V_{\phi^{i}}(s_t))^2] $. 
We will next describe how  semantic tracklets enable  to learn effective VMARL policies.

\section{VMARL with Semantic Tracklets}
\label{sec:app}

Our goal is to develop a representation that permits to effectively learn a multi-agent strategy, \ie, policies $\Pi$, given only visual observations. A good set of policies $\Pi$ maximizes the agents' collaborative abilities to collect  rewards. 

As previously motivated, we aim to mitigate the issue of scalability for VMARL
by designing  intermediate representations for the policy and value networks. For this we study `semantic tracklets,' an object-centric representation that includes  inductive biases about ``where'' and ``what,'' which are useful for VMARL. Semantic tracklets capture both the role and movement of each agent throughout the environment. Given this object-centric representation, we then learn the policy and value functions using graph-nets, as they excel in  tasks that involve reasoning and modeling of interactions. %
In the following we describe our approach using the GFootball environment  as a running example. 

The overall method is illustrated in~\figref{fig:pipeline} and consists of two main components. We highlight the semantic tracklet generation in {\bf \textcolor{gggg}{\contour{gg}{green}}}:  given observations $s_t \in \mathbb{R}^{H\times W \times 3}$ consisting of pixels, we construct the semantic tracklets. Next, highlighted in {\bf \textcolor{bbbb}{\contour{bb}{blue}}}, %
semantic tracklets are transformed into complete graphs, where one object corresponds to a node. Subsequently, policy and value networks predict an action and  the truncated return for each agent. We  describe both components in  detail next.
\begin{figure}[t]
\centering
\includegraphics[width=\linewidth]{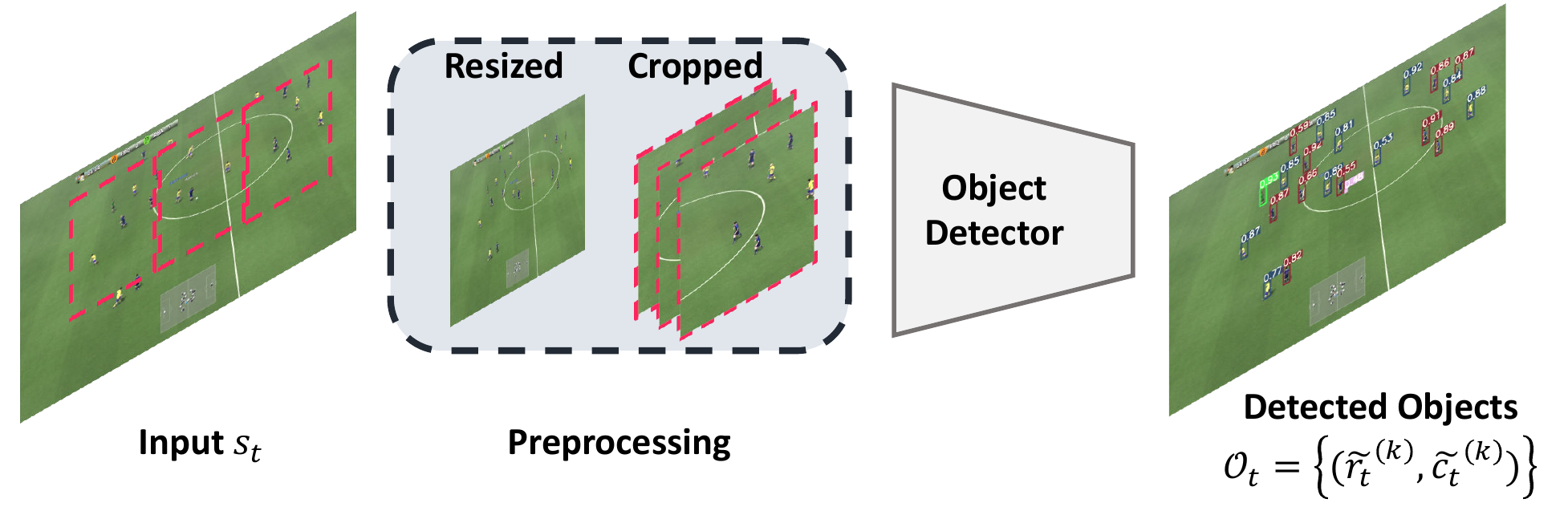}
\caption{Illustration of detection pipeline used for GFootball.
}
\label{fig:det_pipeline}
\vspace{-0.3cm}
\end{figure}

\subsection{Semantic Tracklet Generation}
\label{subsec:tracklet}
Formally, semantic tracklets for a play up to frame $t$ are represented as an \textit{unordered set} of tuples, \ie, 
\be\gX_{\le t} \triangleq \{\rvx^{(1)}_{\le t}, \hdots, \rvx^{(m)}_{\le t}\}.
\ee
The set contains $m$ elements corresponding to the $m$ detected objects, \eg, the players and the ball. Each element is an \textit{ordered tuple}  
\be 
\rvx^{(o)}_{\le t} \triangleq (\rvx^{(o)}_t, \;\rvx^{(o)}_{t-1}, \hdots), o\in\{1, \dots, m\},
\ee 
which is associated with the trajectory of an object, \ie,   $\rvx^{(o)}_{t} \triangleq [\rvr_t^{(o)}, \rvc_t^{(o)}, \rvg_t^{(o)} ]$. Here, $\rvr_t^{(o)}$ represents the object's role, \eg, ball or person in the GFootball environment,  $\rvc_t^{(o)}$ represents each object's spatial-coordinates, and $\rvg_t^{(o)}$ subsumes global information, \eg, distance to the edge of the environment.

To generate the semantic tracklets $\gX_{\le t}$ from an observation $s_t$, we first run an object detector to identify the objects of interest. This process is illustrated in~\figref{fig:det_pipeline}. 
Given the input image $s_t$, we resize and crop the image in preparation for the object detector. This preprocessing permits to detect small objects. %
Subsequently, the object detector yields information about  the relevant objects, \eg, the agents. More details regarding the detectors are  provided in the experimental section.

Formally, the detections are first subsumed in an unordered set of objects 
$\gO_t = \{(\tilde{\rvr}_t^{(k)}, \tilde{\rvc}_t^{(k)})\}$, containing the role and spatial location for the detected objects. We use the tilde symbol (`$\tilde{\cdot}$') to indicate that these detections are not aligned across time, \ie, the $k^{\text{th}}$ object for frames at time $t$ and $t-1$ may differ.

To maintain a consistent correspondence of objects across time, we will align the objects in $\gO_t$. For this we use an object tracking formulation 
which identifies the correspondence between objects in $\gO_{t-1}$ and $\gO_{t}$.

Specifically, assume we are given an ordered set of objects $\{(\rvr_{t-1}^{(o)}, \rvc_{t-1}^{(o)})\}$ for the previous time step $t-1$. %
Intuitively, tracking of the ``unordered'' objects
$\gO_t = \{(\tilde{\rvr}_t^{(k)}, \tilde{\rvc}_t^{(k)})\}$ is equivalent to assigning each element in the unordered set %
to an element from the ordered set %
of the previous time step. 
This is formulated as the unbalanced assignment problem 
\bea\nonumber
\min_{\mM_t} \sum_{o,k} \mM_{t, ok} \left( \norm{\rvc_{t-1}^{(o)}-\tilde{\rvc}_t^{(k)}}_2^2 + d(\rvr_{t-1}^{(o)},\tilde{\rvr}_t^{(k)}) \right),\\
\label{eq:tracking}
\;\;\text{s.t.}\; \mM_{t, ok} \in \{0,1\}, \sum_{o}\mM_{t, ok}\leq1, \sum_{k}\mM_{t, ok}=1.
\eea
Hereby $\mM_t$ denotes an assignment matrix. The cost for assigning $\tilde{\rvc}_t^{(k)}$ to ${\rvc}_{t-1}^{(o)}$ is the sum of an $\ell_2$-loss on the object's coordinates and a distance  $d$ between their roles. For example, when $\rvr_t^{(o)}$ is categorical, $d$ can be the zero-one loss. This assignment cost encourages to match objects with the same role that are spatially close.%

We solve the program given in~\equref{eq:tracking} by reducing the unbalanced assignment problem to a balanced one before using the Hungarian algorithm~\cite{kuhn1955hungarian}. This is achieved by adding surrogate matches with zero loss. Given the assignment $\mM_t$, we update the location estimates to prepare for the next time step, \ie, $\forall o$  $(\rvr_t^{(o)},\rvc_t^{(o)}) \leftarrow (\tilde{\rvr}_t^{(k)}, \tilde{\rvc}_t^{(k)})$ %
if object $k$ is assigned to object $o$, \ie, if $\mM_{t, ok} = 1$.
If no object is assigned, \ie, if it is assigned to a surrogate match, we keep the previous tracked roles and locations, \ie, $(\rvr_t^{(o)},\rvc_t^{(o)}) \leftarrow (\rvr_{t-1}^{(o)},\rvc_{t-1}^{(o)})$. This case happens when there are missing detections.

With the objects aligned across time, we then compute the global information $\rvg_t^{(o)}$ to complete our proposed semantic tracklets $\gX_{\le t}$. 
We will next describe how to use graph-nets to learn policies and value functions from this object-centric representation.

\begin{table*}[t]
\centering
\small
\setlength{\tabcolsep}{12pt}
\begin{tabularx}{\textwidth}{ccc@{\hspace{0.5cm}}c@{\hspace{0.5cm}}c@{\hspace{0.5cm}}c@{\hspace{0.5cm}}c@{\hspace{0.5cm}}c}
\specialrule{.15em}{.05em}{.05em}
\multirow{2}{*}{Inter. Rep.} & \multirow{2}{*}{Arch.} & \multicolumn{2}{c}{\makecell{\em Visual Cooperative Navigation}} & \multicolumn{2}{c}{\makecell{\em Visual Prey and Predator}} & \multicolumn{2}{c}{\makecell{\em Visual Cooperative Push}}\\
                                           &                                    &    $N=3$&  $N=6$ &  $N=3$ & $N=6$                                                   & $N=3$ &$N=6$ \\
\hline
None & CNN & -654.7$\pm21.8$& -4780.4$\pm$272.1 & -27.3$\pm$0.4& -107.2$\pm$1.2& -349.5$\pm$1.1& -1384.1$\pm$32.2\\
Segmentation & CNN & -653.1$\pm$4.6 & -4518.0$\pm$226.1 & -26.3$\pm$0.4 & -96.6$\pm$2.4 & -352.9$\pm$0.9 & -1494.3$\pm$62.1\\
Tracklets & MLP  &  -392$\pm$3.1  & -4040.0$\pm$12.4 &  -1.3$\pm$1.4 &  46.8$\pm$3.4 & -345.4$\pm$1.4 &  -1360.9$\pm$29.5\\
Tracklets & GCN &\bf{ -381.1$\pm$11.2} & \bf{-3642.7$\pm$46.8} & \bf{3.6$\pm$2.1} & \bf{279.4$\pm$24.3} & \bf{-217.6$\pm$19.5} & \bf{-1098.9$\pm$70.2}\\
\specialrule{.15em}{.05em}{.05em}
\end{tabularx}
\vspace{-0.4cm}
\caption{Average evaluation episode rewards of baselines and our approach on VMPE tasks.
}
\label{tab:quan_results_mpe}
\vspace{-0.65cm}
\end{table*}

\subsection{Policy and Value Networks}
\label{sec:policy}

Classical deep nets operate on a vector or matrix. Concatenating all elements in the tracklet $\gX_{\le t}$ is an intuitive way to construct a vector.
However, vectorizing \textit{implicitly} assumes an ordering of the set's elements. Note, permuting the elements in the input vector of a standard deep net will result in a \textit{different} deep net output. This is undesirable as the environment configuration did not change when permuting the agents. Therefore, we  use a graph-net to model the policy $\pi_{\theta^i}(\cdot|\gX_{\le t})$ and the value function $V_{\phi^i}(\gX_{\le t})$. %
This guarantees permutation invariance \wrt  the agents' ordering, \ie, the graph-net's output remains identical irrespective of the input permutation. Specifically, we use graph convolutional nets (GCNs) to model both functions. 

\noindent\textbf{Graph Construction from Semantic Tracklets.}
For each controlled agent, we construct a complete graph using the $K$ nearest neighbor objects. Each object $o$ %
is a node. For each node $o$, we use an embedding $\Phi_{0}^{(o)}$, %
 constructed from the semantic tracklets.  Each node embedding uses the most recent four frames of the semantic tracklets, \ie, $\Phi_0^{(o)} \triangleq [\rvx^{(o)}_t, \rvx^{(o)}_{t-1}, \rvx^{(o)}_{t-2}, \rvx^{(o)}_{t-3}]$. 
We next describe the GCN  architecture to learn the policy and value functions from the node embeddings.

\noindent\textbf{Permutation Invariant Architecture.}
A graph convolution $f_{\text{g}}^{(l)}$ at layer $l$ is defined as follows: 
\begin{equation}
f_{\text{g}}^{(l)} (\Phi_{l-1}) \triangleq  \sigma \left(
\frac{\mA\Phi_{l-1}W_{\text{other}}^{(l)} +  \Phi_{l-1}W_{\text{self}}^{(l)}}{K+1}\right).
\end{equation}
Here, $K+1$ denotes the number of nodes in the graph, \ie, the $K$ nearest neighbors plus the agent's node. We let $\mA$ denote the adjacency matrix of the graph, $\Phi_{l} \in \mathbb{R}^{(K+1)\times K_{\text{in}}^{(l)}}$ denotes the feature matrix at the $l^{\text{th}}$ layer, where each row is a $K_{\text{in}}^{(l)}$-dimensional feature $\Phi_l^{(i)}$ of an object. The layer's trainable parameters are $W_{\text{other}}^{(l)}, W_{\text{self}}^{(l)} \in \mathbb{R}^{K_{\text{in}}^{(l)} \times K_{\text{out}}^{(l)}}$, {where
$K_{\text{in}}^{(l)}$ and  $K_{\text{out}}^{(l)}$ denote  the input and output dimensions of the $l^{\text{th}}$ layer.} %

Our policy and value network share parameters for their GCN. However,  their fully connected layers, \ie, their multi-layer perceptron (MLP), differ. This gives rise to the following formulation: 
\bea\label{eq:pool}
\Phi_{L}  &\!\!=&\!\! \text{MaxPool}\left(f_g^{L} \circ f_g^{L-1} \circ \hdots \circ f_g^{1}(\Phi_0) \right)\\
\label{eq:pn}
\pi_{\theta^i}(\cdot|\rvx_{\le t}) &\!\!=&\!\! \text{Softmax}(f_{\pi, \text{MLP}}(\Phi_{L}))\\
\label{eq:vn}
V_{\phi^i} (\rvx_{\le t}) &\!\!=&\!\! f_{V,\text{MLP}}(\Phi_{L}).
\eea 
Note that each row of $\Phi_0$ consists of semantic tracklets $\Phi_{0}^{(o)}$ of object $o$. %
The max pooling, in~\equref{eq:pool}, is performed over the first (agent) dimension. This ensures a permutation invariant representation  $\Phi_{L}$ as max-pooling  ignores the permutation.
 In~\equref{eq:pn}, $f_{\pi,\text{MLP}}$ %
refers to 
a MLP to model the policy function.
 Similarly, $f_{V,\text{MLP}}$ in ~\equref{eq:vn} %
refers to a MLP
for the value network. These MLPs consist of fully connected layers with ReLU non-linearity. To learn the policy functions' parameters $\{\theta^i\}_{i=1}^N$, %
and the value functions' parameters $\{\phi^i\}_{i=1}^N$, %
the PPO algorithm described in~\secref{sec:prelim} is used.

\section{Experiments}
\label{sec:exp}

We conduct experiments on two visual multi-agent environments: the Visual Multiple Particle Environment (VMPE), which mimics the original MPE~\cite{maddpg, pic}, and the Google Research Football (GFootball) environment. %
We demonstrate the success of `semantic tracklets' in those visual multi-agent environments,  controlling up to $6$ visual agents on the same team at once.

\subsection{Visual MPE}
\vspace{-0.1cm}
{\noindent \bf Environment details.}
In VMPE, the observation of each agent is  an image with agents and landmarks  rendered in different colors. %
This differs from the original MPE~\cite{maddpg, pic}, where the observation of each agent is a compact vector which summarizes the information of the environment, \ie, the location and velocity of other agents. In the VMPE  we consider  three tasks with $N=3$ and $N=6$ agents: %

{\em Visual Cooperative Navigation:} $N$ agents work cooperatively to cover $N$ landmarks. The environment reward encourages agents to cover all landmarks.

{\em Visual Prey and Predator:} $N$ predators work together to capture $N/3$ faster moving preys. The predators receive positive rewards when colliding with a prey and receive negative rewards when colliding with fellow predators. 

{\em Visual Cooperative Push:} $N$ agents work cooperatively to push a large heavy ball to a landmark. The agents are rewarded when the big ball approaches the landmark.

\begin{figure}[t]
\centering
\renewcommand{\arraystretch}{0.5}
\setlength{\tabcolsep}{1pt}
\begin{tabular}{cccc}
\includegraphics[width=0.24\linewidth, ]{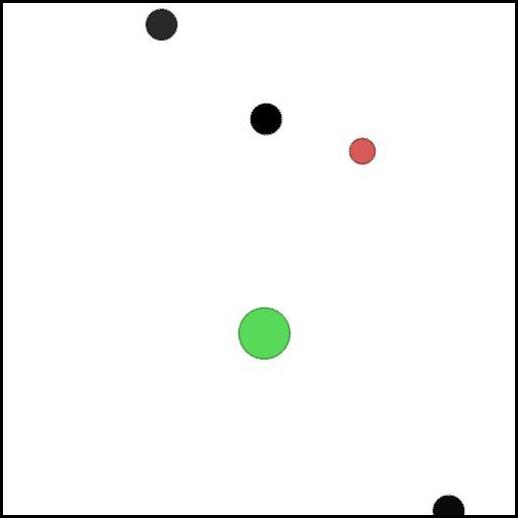}&
\includegraphics[width=0.24\linewidth, ]{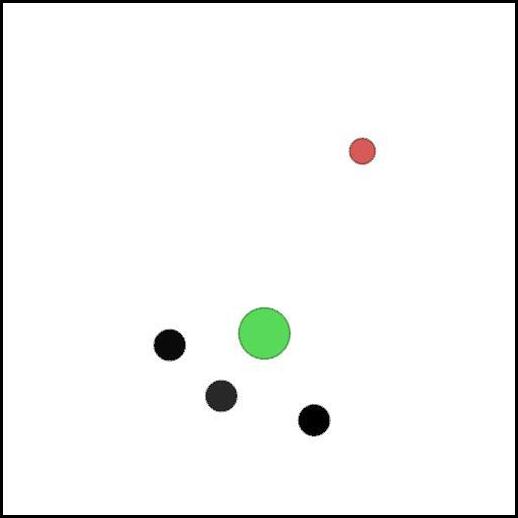}&
\includegraphics[width=0.24\linewidth, ]{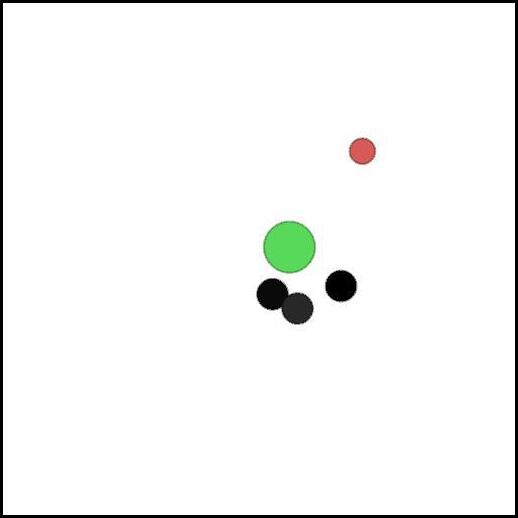}&
\includegraphics[width=0.24\linewidth, ]{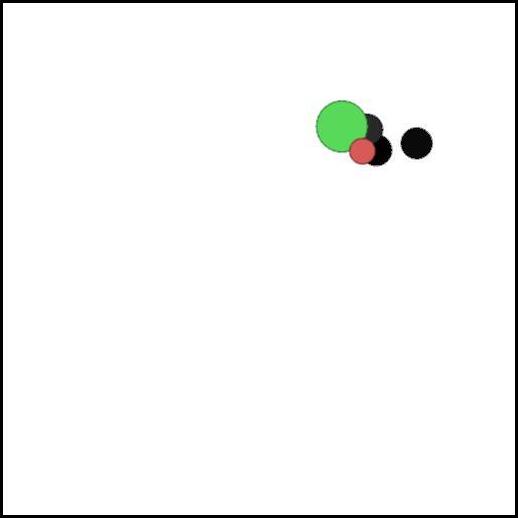}\\
\scriptsize(a) & \scriptsize(b)& \scriptsize(c)& \scriptsize(d)
\end{tabular}
\caption{Qualitative results for {\em Visual Cooperative Push} ($N=3$) using semantic tracklets. We observe 
agents working together to push the  large green ball to the red landmark.
}
\label{fig:qual_result_vmpe}
\vspace{-0.5cm}
\end{figure}

\begin{figure*}[t]
\centering
\begin{minipage}{.4\textwidth}
\includegraphics[width=1\linewidth]{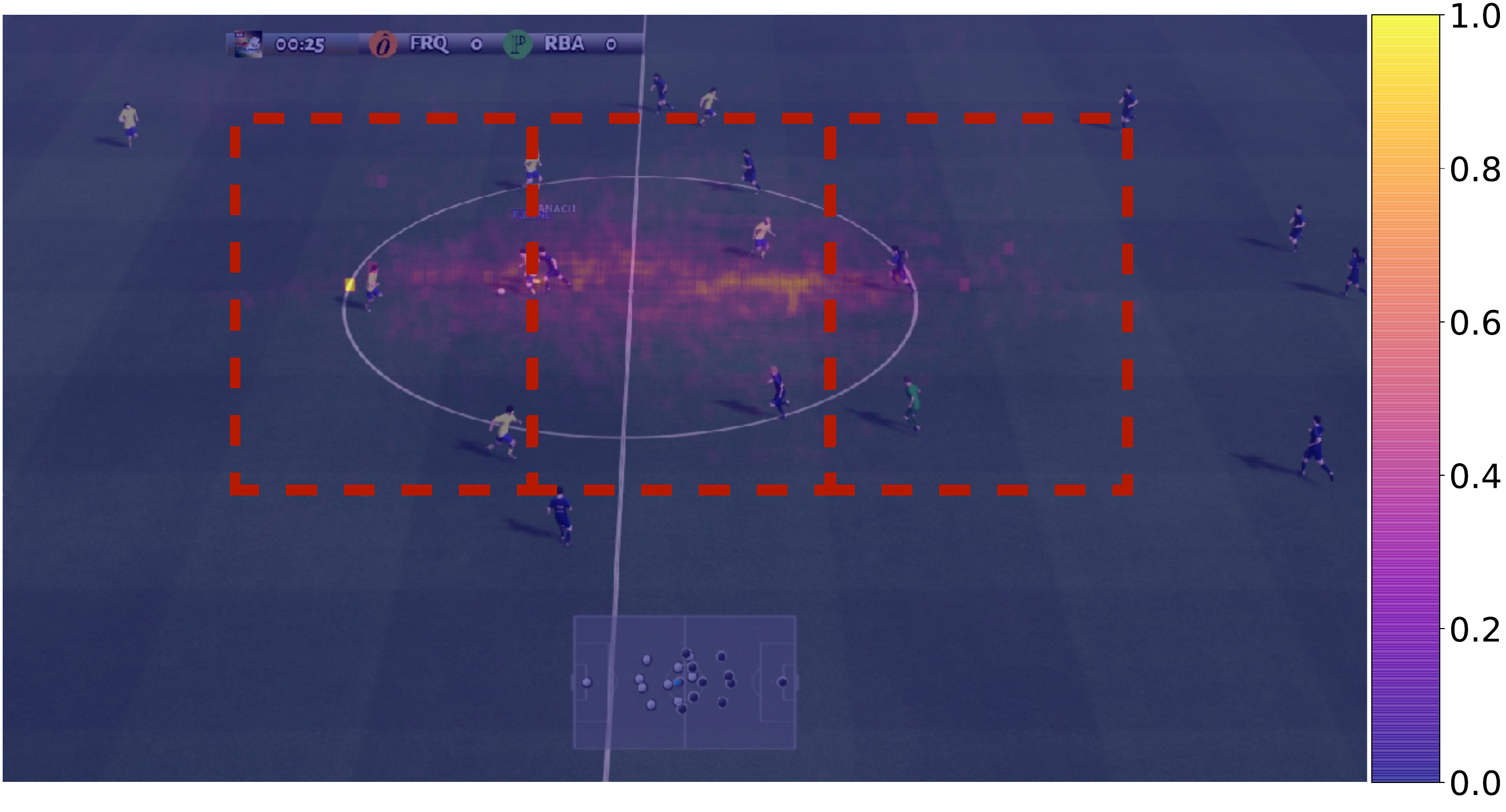}
\caption{Probability of ball location in the camera coordinate system. 
As expected, the camera follows the ball. Hence, the ball appears more likely close to the image center. 
}
\label{fig:ball_heatmap}
\end{minipage}
\hspace{.2em}
\begin{minipage}{.5\textwidth}
\setlength{\tabcolsep}{0pt}
\begin{tabular}{c}
\includegraphics[width=1\linewidth, trim={0 0 0 3.2cm},clip]{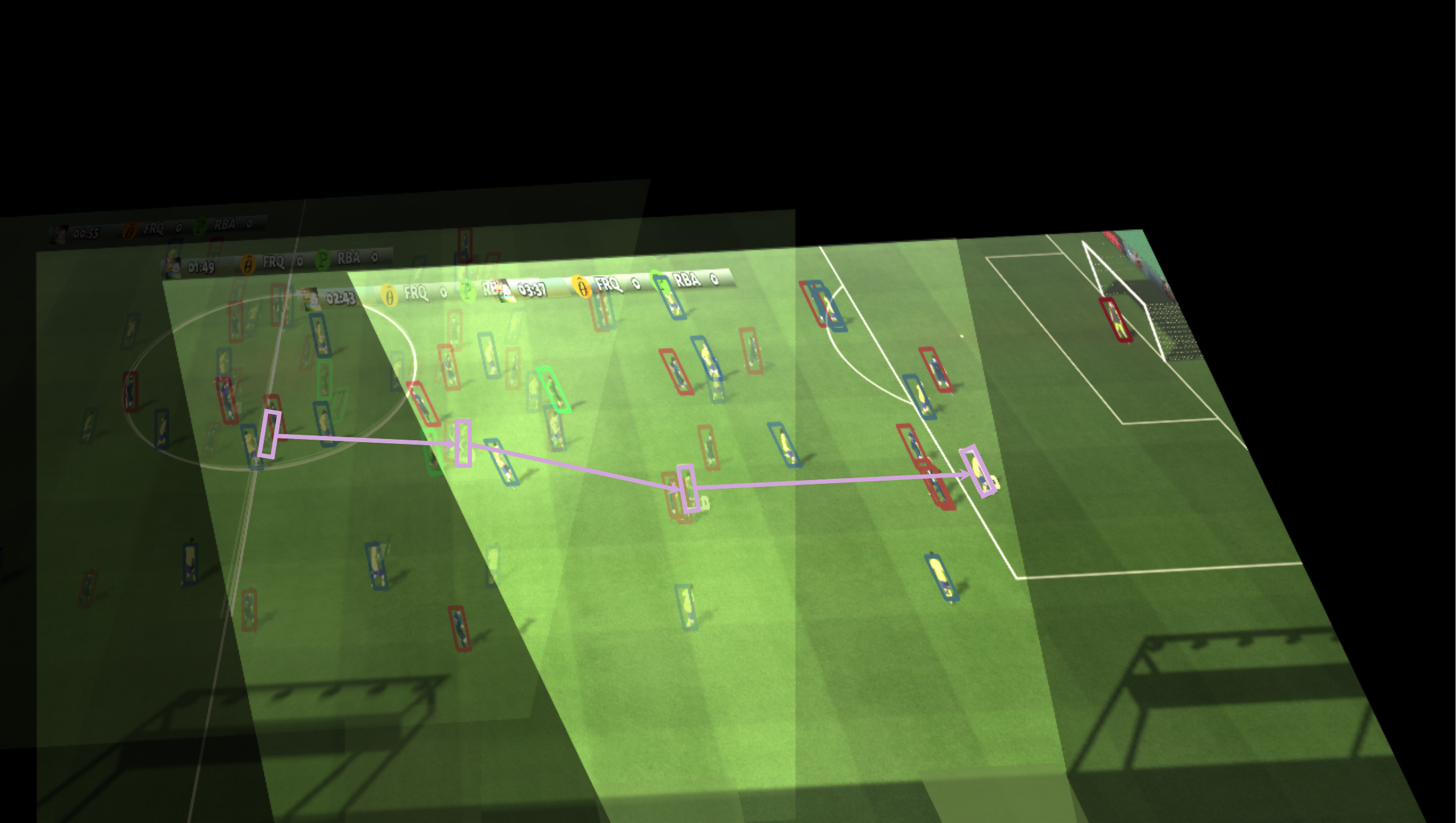}\\
\end{tabular}
\caption{
Visualization of the tracking results across different frames with a moving camera. As can be seen, our tracking successfully maintains the identity of the active player, boxed in {\bf \color{pp} purple}, across frames.
}
\label{fig:tracking_stitch_field}
\end{minipage}
\vspace{-0.0cm}
\end{figure*}
{\noindent \bf Baselines and metrics.}
We consider a baseline without an intermediate representation, \ie,  pixels are directly passed to the policy and value networks. We refer to this via None+CNN.  Inspired by Hong \etal~\cite{Hong17}, we also compare with  semantic segmentation as an intermediate representation, which we refer to via Segmentation+CNN. 
To avoid semantic segmentation errors undermining the baseline's performance, \ie, for a stronger baseline, we use ground truth segmentations to train and test the baseline.

For both of the aforementioned representations, the policy and value networks are CNNs. 
To evaluate our approach, we run $100$ evaluation episodes every $1,000$ training episodes.
To ensure that the evaluation is rigorous and fair, we follow the evaluation protocol suggested by Colas \etal~\cite{Colas18} and Henderson \etal~\cite{Henderson17}.
We report the final metric, which is the average reward over the last $1,000$ evaluation episodes, \ie, $100$ episodes for each of the last ten policies during training. %

{\noindent \bf Implementation details.}
For semantic tracklets, we use simple thresholding techniques to detect the locations of landmarks and agents. %
The detected locations and roles are used to construct semantic tracklets. 
Following~\cite{maddpg}, we use multi-agent deep deterministic policy gradient (MADDPG) to train our approach and all baselines. Agents are trained for $60, 000$ episodes in all tasks (episode length is either 25 or 50 steps). %

{\noindent \bf Visual multi-agent results.}
In~\tabref{tab:quan_results_mpe}, we report quantitative results in final metrics.
Across tasks and for different numbers of visual agents we observe  semantic tracklets to consistently outperform the baselines that directly use pixels or semantic segmentation as intermediate representation. %
We also observe that the improvement of using semantic segmentation as intermediate representation over directly using pixels is marginal in this environment. 
Moreover, semantic tracklets with GCN architecture achieve better rewards than semantic tracklets with MLP. This demonstrates the effectiveness of a GCN-based policy and value network. %
We visualize our learned polices on {\em Visual Cooperative Push} ($N=3$) in \figref{fig:qual_result_vmpe}, where agents  successfully coordinate their behavior to push the large green ball to the red landmark.%

{\noindent \bf Robustness to tracklet quality.} We further investigate how tracklet quality impacts RL results. %
We consider randomly dropping objects' information, \ie, location and role, at different rates %
{using {\em Visual Cooperative Navigation} $(N=3)$, {\em Visual Prey and Predator} $(N=6)$, and {\em Visual Cooperative Push} $(N=3)$. {%
The results are summarized in~\tabref{tab:quan_mpe_dropout}. As shown in~\tabref{tab:quan_mpe_dropout}, when $10\%$ of information is dropped, the rewards drop only slightly across different tasks. %
When we drop $40\%$ of the information, the reward drops around $10.7\%$. }}
\begin{table}[t]
\centering
\small
\setlength{\tabcolsep}{2.8pt}
\begin{tabularx}{1.03\linewidth}{c|cccc}
\specialrule{.15em}{.05em}{.05em}
Dropout rate ($\%$)& $0\%$ & $10\%$ &$20\%$&$40\%$  \\
\hline
\makecell{\emph{Visual Coop.}\\ \emph{Navigation ($N=3$)}}& -381.1$\pm$11 & -382.2$\pm$9  & -391.2$\pm$4 & -409.1$\pm$1\\
\hline
\makecell{\emph{Visual Prey $\&$}\\ \emph{Pradator ($N=6$)}}& 279.4$\pm$24 & 273.5$\pm$4  & 253.6$\pm$10 & 252.2$\pm$2\\
\hline
\makecell{\emph{Visual Coop.}\\ \emph{Push ($N=3$)} }&-217.6$\pm$19 & -225.3$\pm$10  & -233.4$\pm$27 & -250.3$\pm$6\\
\specialrule{.15em}{.05em}{.05em}
\end{tabularx}
\vspace{-0.4cm}
\caption{Semantic tracklets' average evaluation rewards \vs different dropout rates.
}
\label{tab:quan_mpe_dropout}
\vspace{-0.9cm}
\end{table}

\begin{table*}[t]
\centering
\small
\setlength{\tabcolsep}{9pt}
\begin{tabularx}{1\textwidth}{cccccccccc}
\specialrule{.15em}{.05em}{.05em}
Inter. Rep. & Arch. & \makecell{ {\em 11 $\vs$ 11} \\($N=3$)} &\makecell{ {\em 11 $\vs$ 11} \\($N=5$) }& \makecell{{\em single goal \vs lazy} \\($N=3$)} & \makecell{{\em 3 \vs 1} \\ with keeper \\($N=3$)} & \makecell{{\em run, pass and} \\shoot with keeper\\ ($N=2$)} & \makecell{pass and shoot \\with keeper\\ ($N=2$)} \\
\hline
None~\cite{gfootball} & CNN & -1.50$\pm$0.7  & -0.75$\pm$0.5 & 0.10$\pm$0.1 & 0.05$\pm$0.07 & 0.03$\pm$0.06  & 0.08 $\pm$0.05  \\
Tracklets & GCN & \bf{0.54$\pm$1.3} & \bf{1.67$\pm$1.8} & \bf{0.75$\pm$0.3} & \bf{0.98$\pm$0.07} & \bf{0.70$\pm$0.11} & \bf{ 0.94$\pm$0.02}\\
\specialrule{.15em}{.05em}{.05em}
\end{tabularx}
\vspace{-0.4cm}
\caption{Average score difference of baselines and our approach on multi-agent visual GFootball tasks. 
}
\label{tab:quan_results_football}
\vspace{-0.8cm}
\end{table*}
\begin{figure}[t]
\centering
\includegraphics[width=0.512\linewidth]{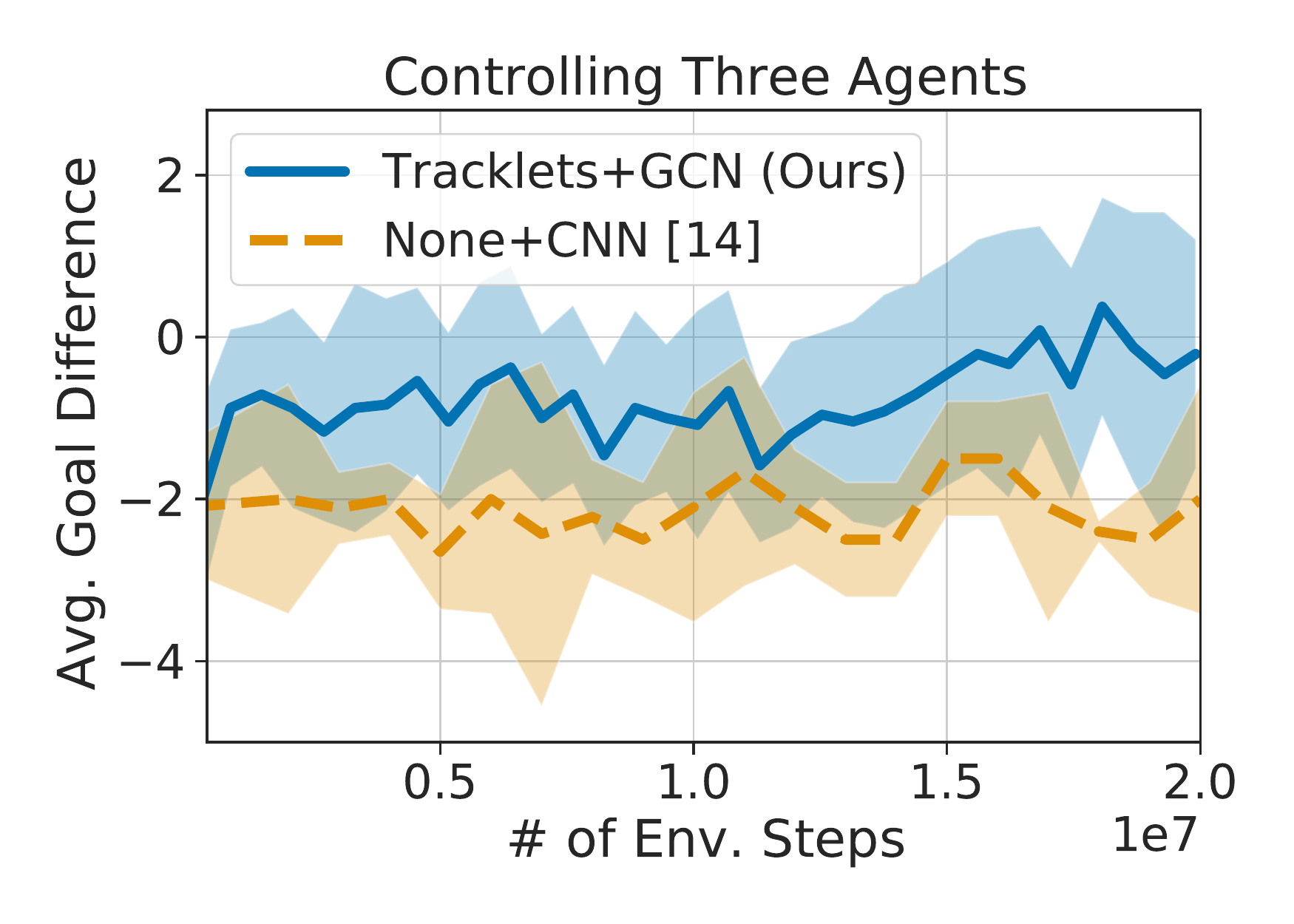}
\hspace{-0.45cm}
\includegraphics[width=0.512\linewidth]{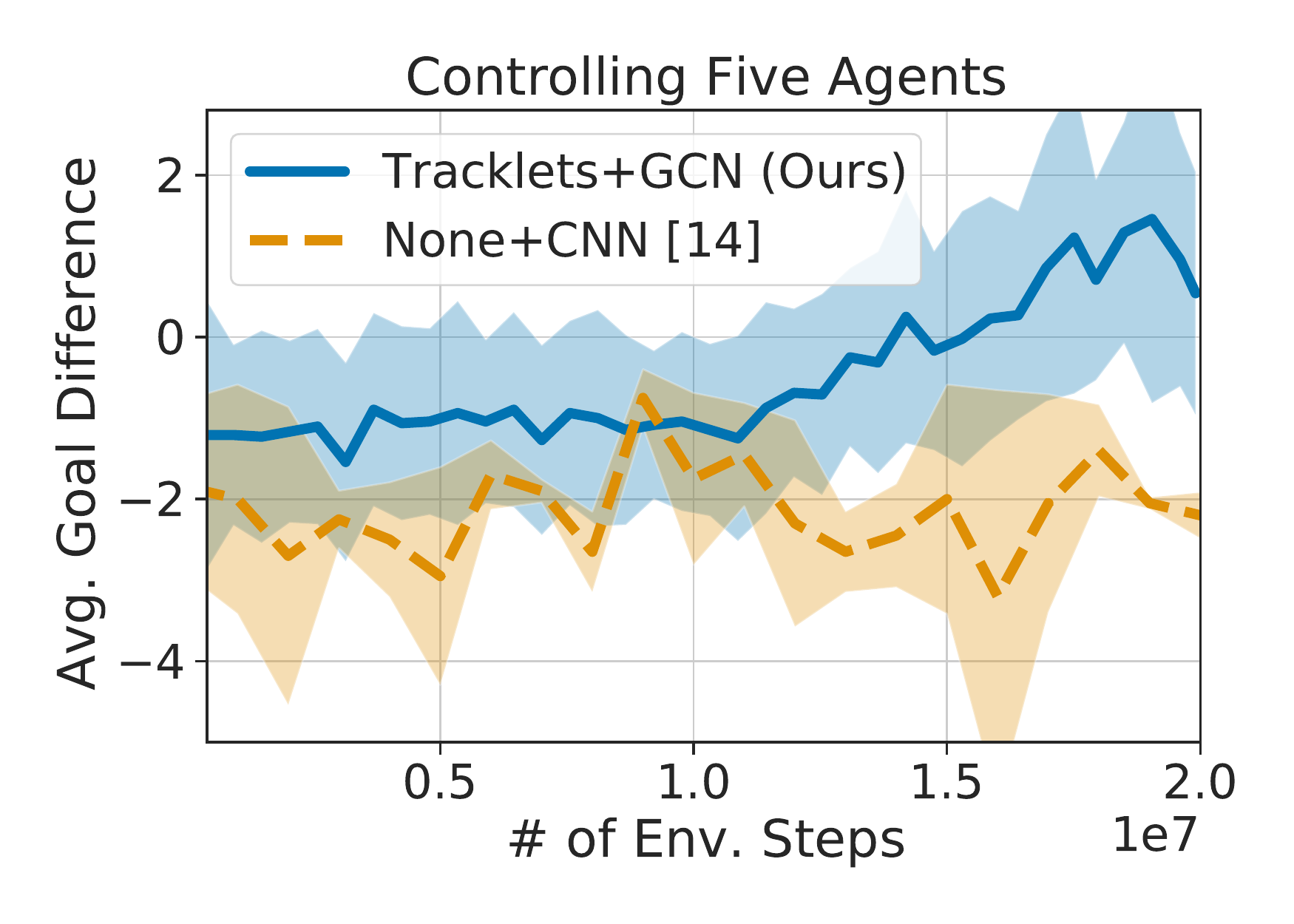}
\vspace{-0.9cm}
\caption{Training curves. {\bf{Left:}} Controlling three agents. {\bf{Right:}} Controlling five agents. 
} 
\vspace{-0.7cm}
\label{fig:quan_results_football}
\end{figure}

\begin{figure*}[t]
\centering
\setlength{\tabcolsep}{3pt}
\renewcommand{\arraystretch}{0.6}
\begin{tabular}{ccc}
\includegraphics[width=0.28\textwidth, trim={0 5.5cm 7cm 0},clip]{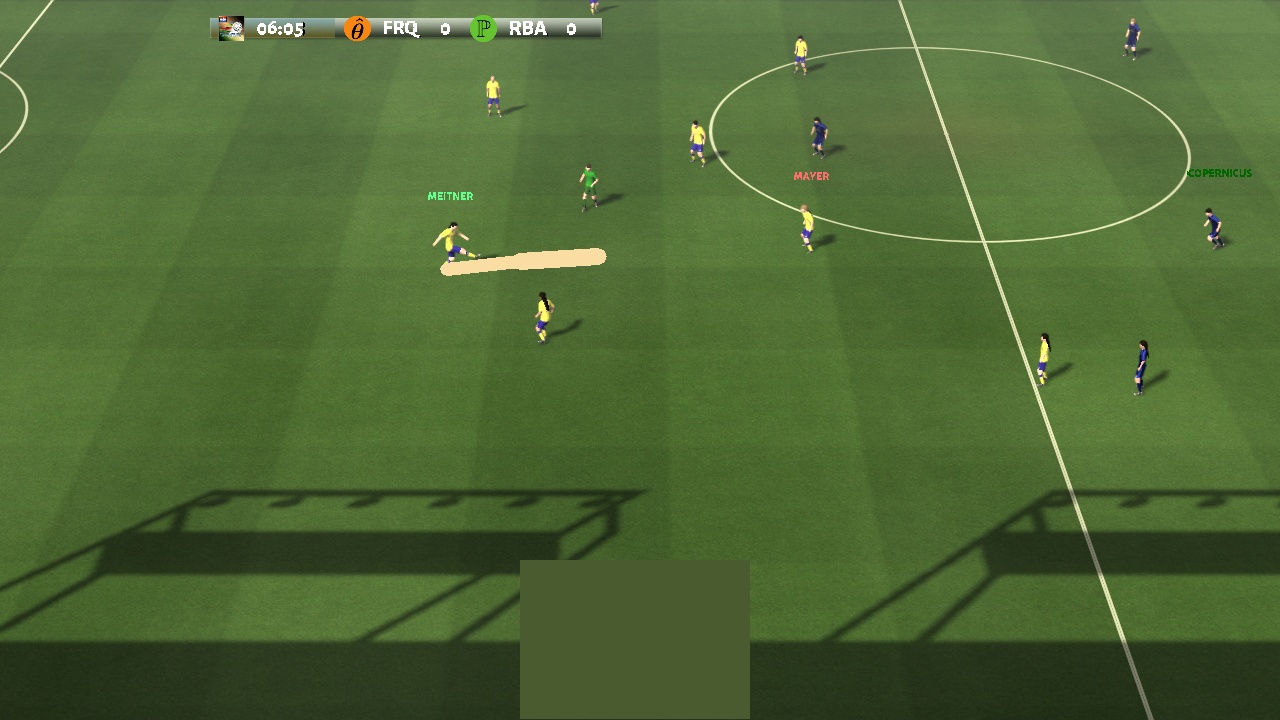}&
\includegraphics[width=0.28\textwidth, trim={0 5.5cm 7cm 0},clip]{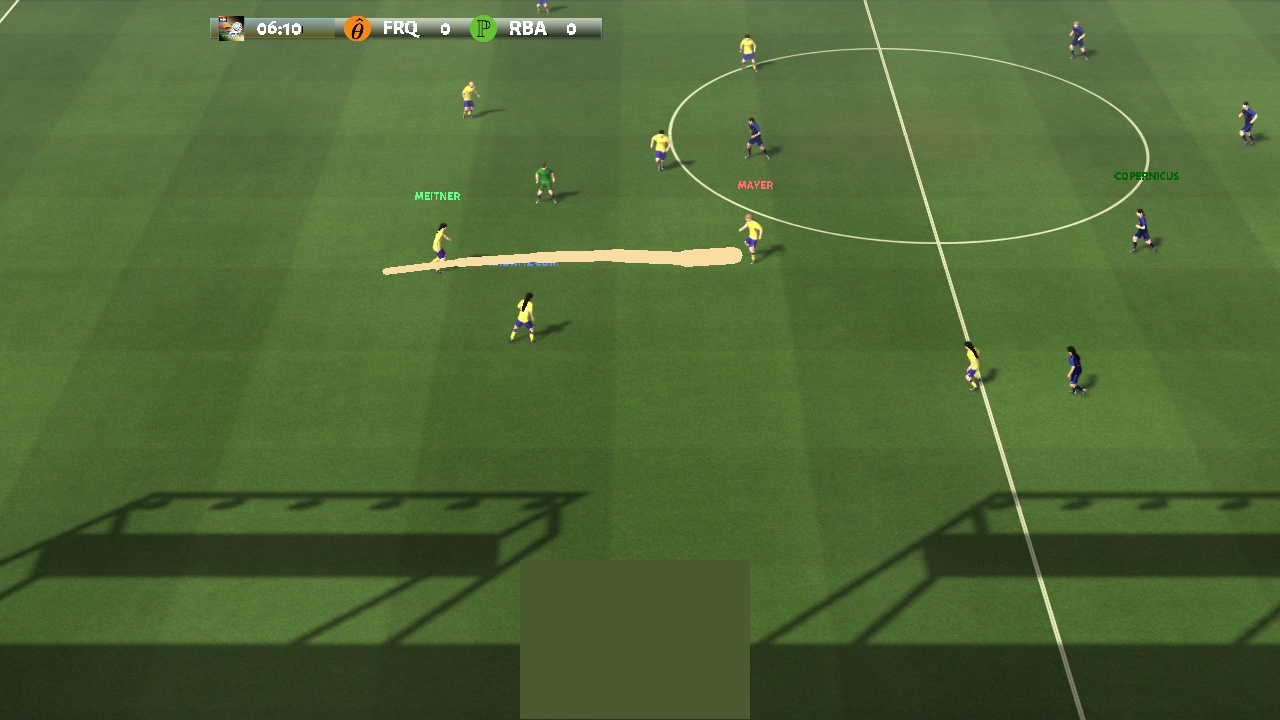}&
\includegraphics[width=0.28\textwidth, trim={0 5.5cm 7cm 0},clip]{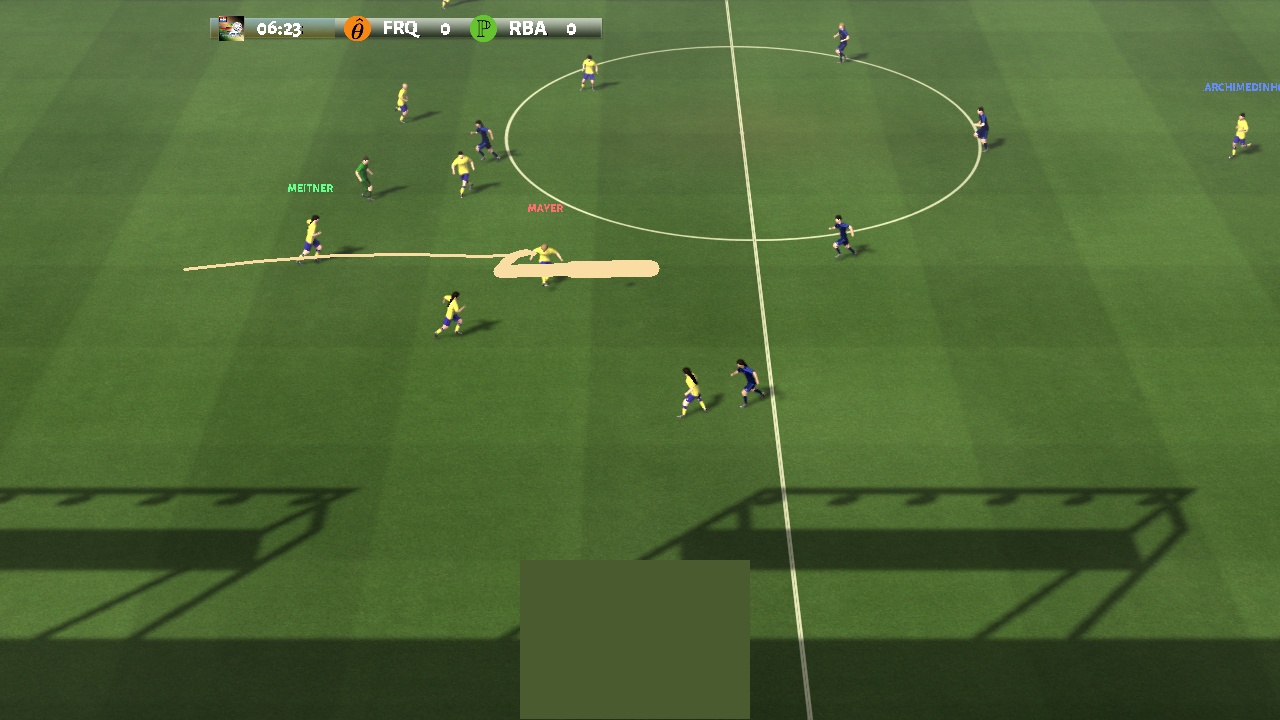}\\
\scriptsize(a) & \scriptsize(b) & \scriptsize(c)\\
\includegraphics[width=0.28\textwidth, trim={0 5.5cm 7cm 0},clip]{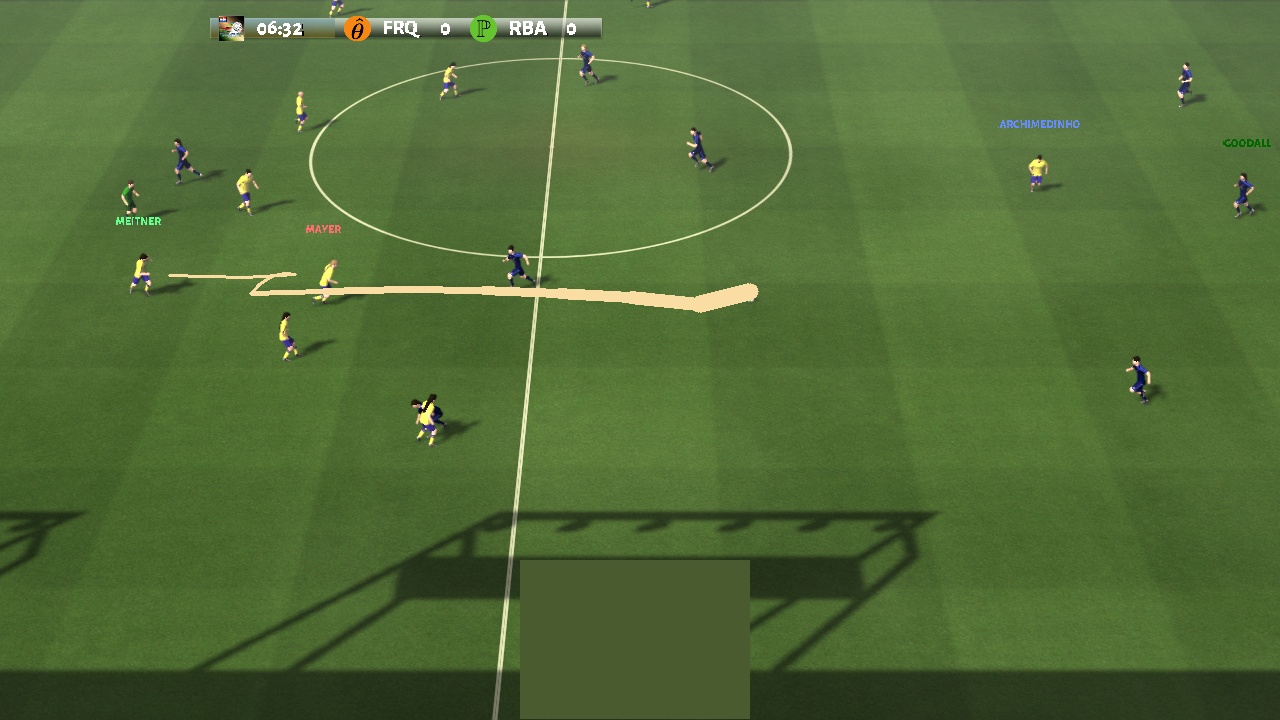}&
\includegraphics[width=0.28\textwidth, trim={0 5.5cm 7cm 0},clip]{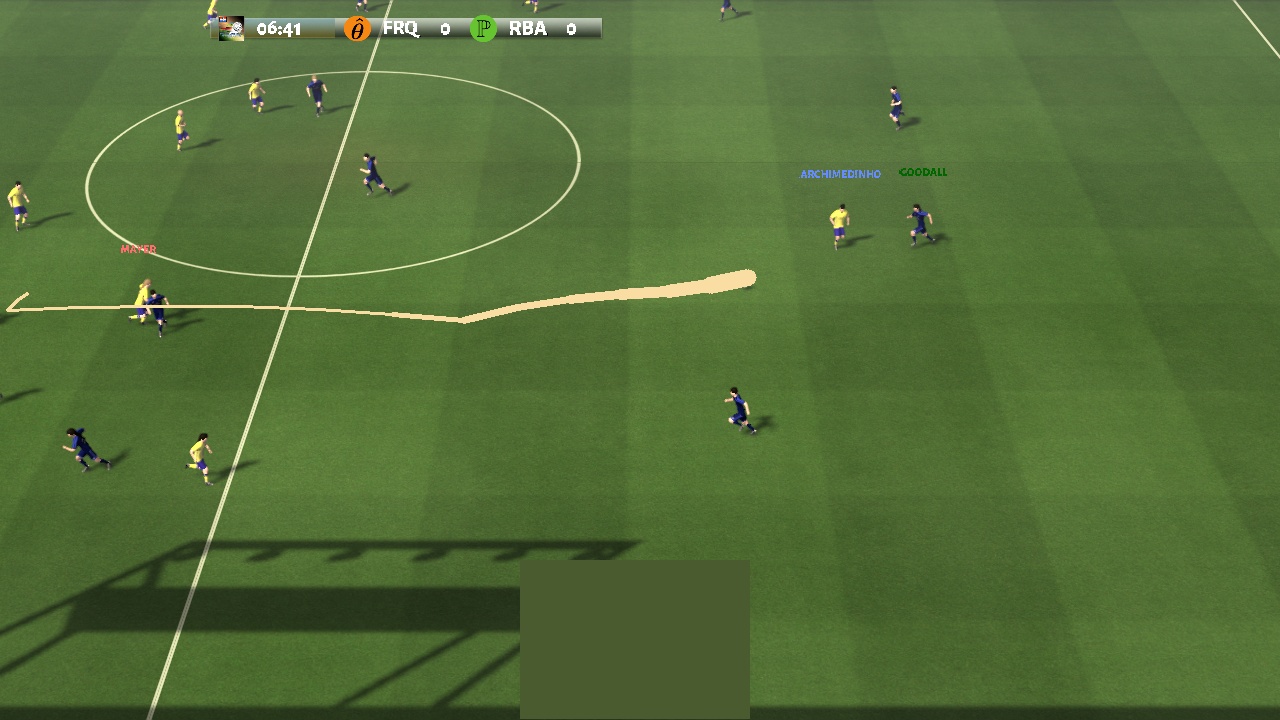}&
\includegraphics[width=0.28\textwidth, trim={0 5.5cm 7cm 0},clip]{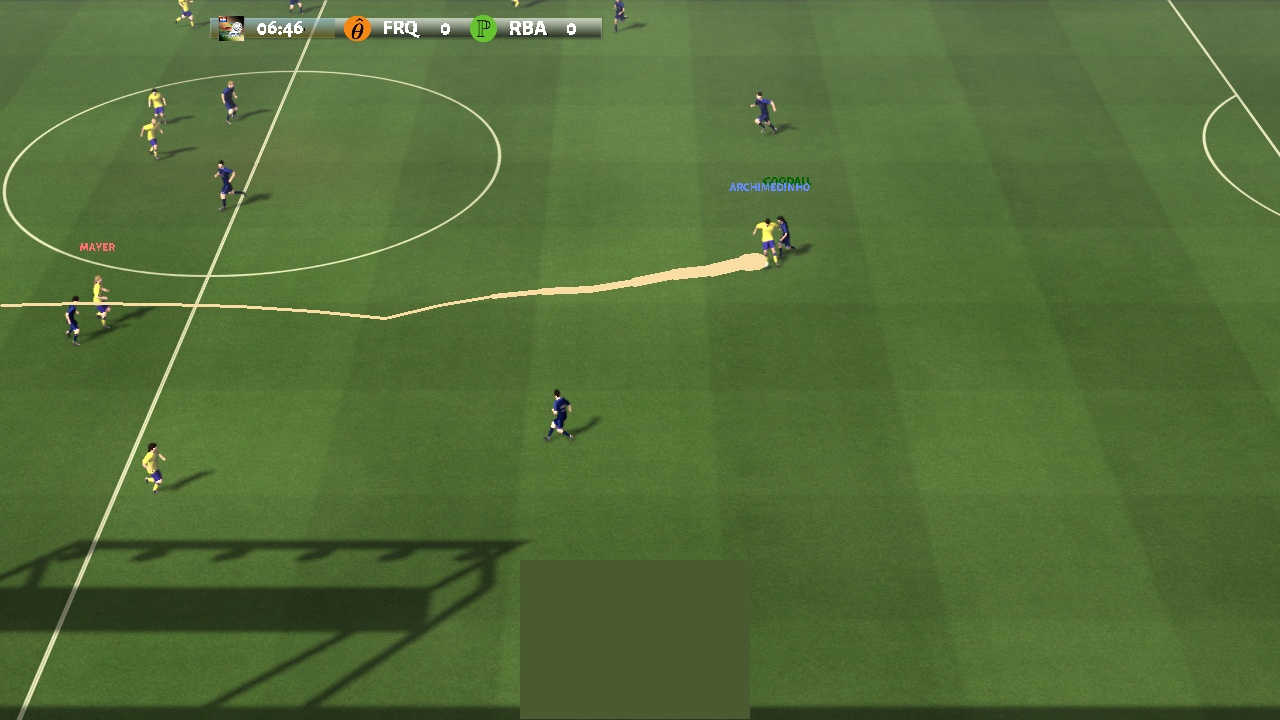}\\
\vspace{-0.0cm}
\scriptsize(d) & \scriptsize(e) & \scriptsize(f)\\
\specialrule{.2em}{.05em}{.05em}\vspace{+0.0cm}\\
\includegraphics[width=0.28\textwidth, trim={0 5.5cm 7cm 0},clip]{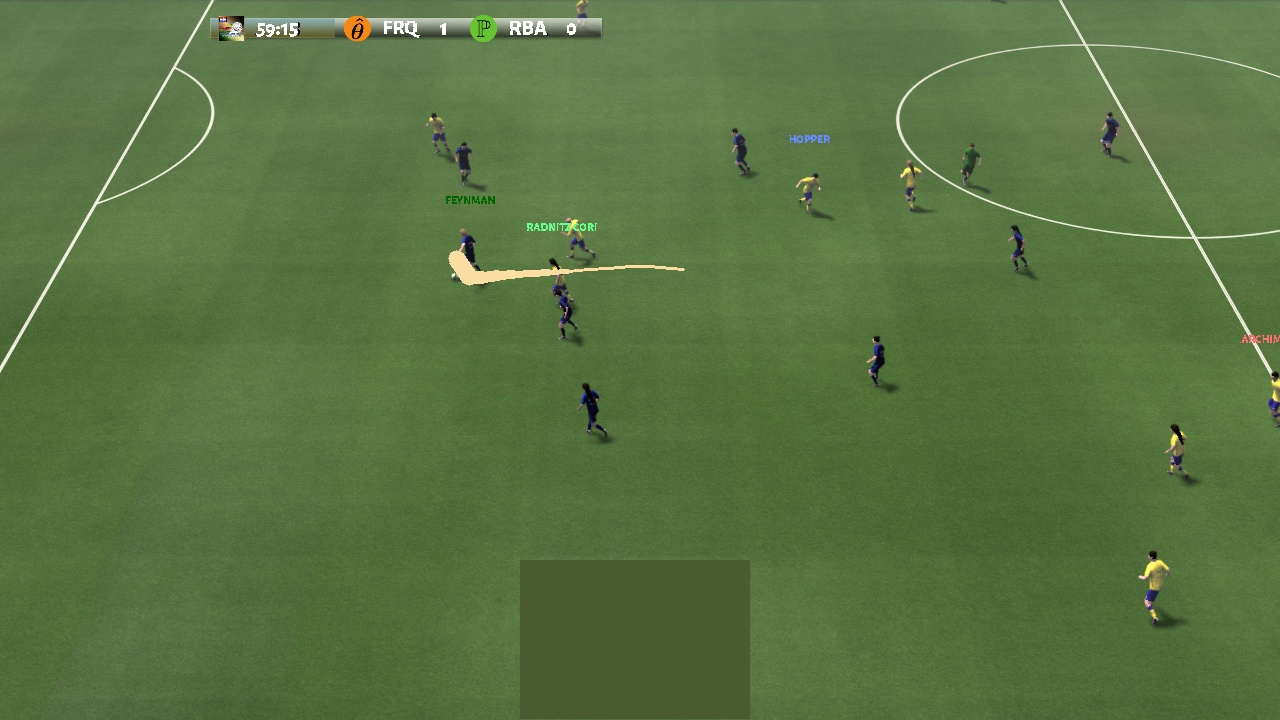}&
\includegraphics[width=0.28\textwidth, trim={0 5.5cm 7cm 0},clip]{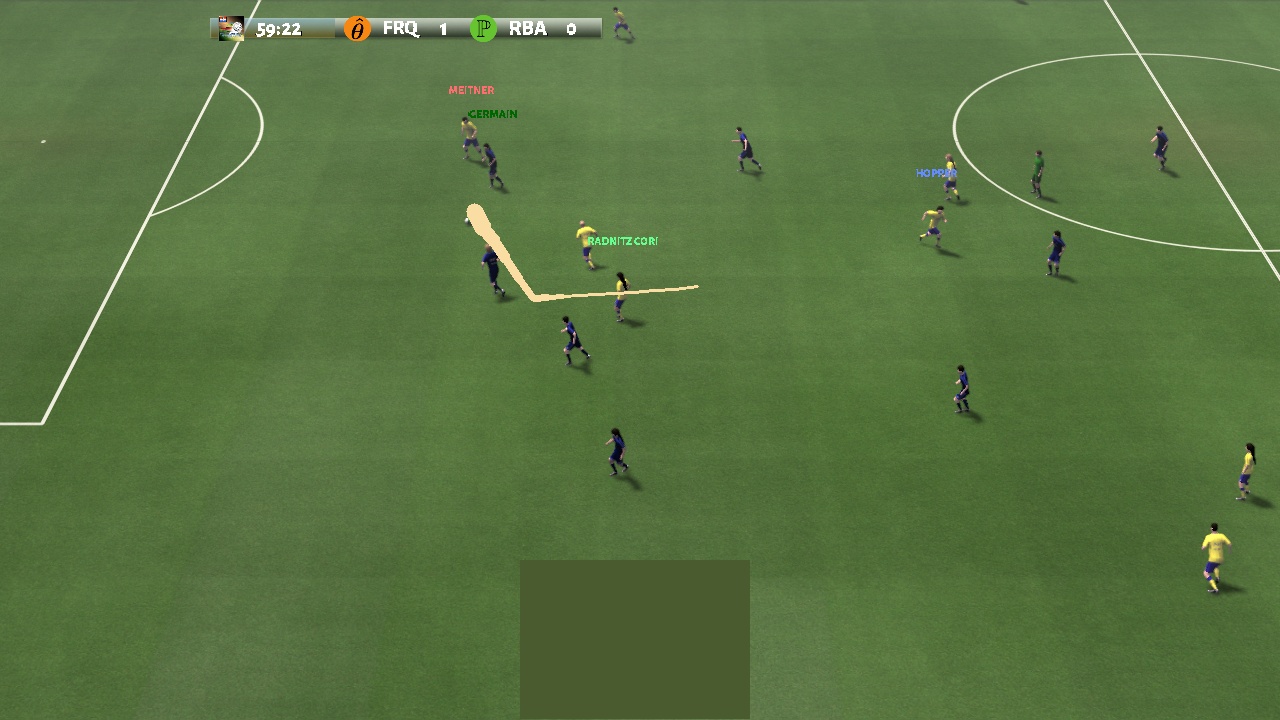}&
\includegraphics[width=0.28\textwidth, trim={0 5.5cm 7cm 0},clip]{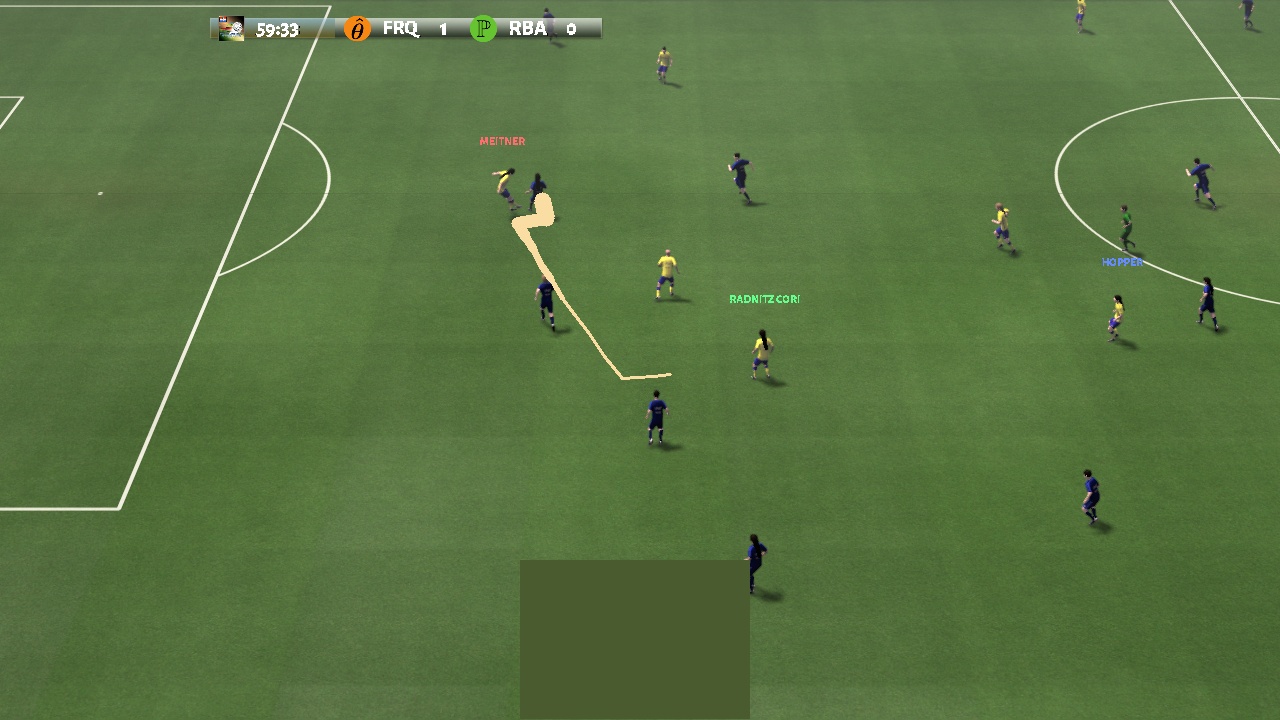}\\
\scriptsize(g) & \scriptsize(h) & \scriptsize(i)\\
\includegraphics[width=0.28\textwidth, trim={0 5.5cm 7cm 0},clip]{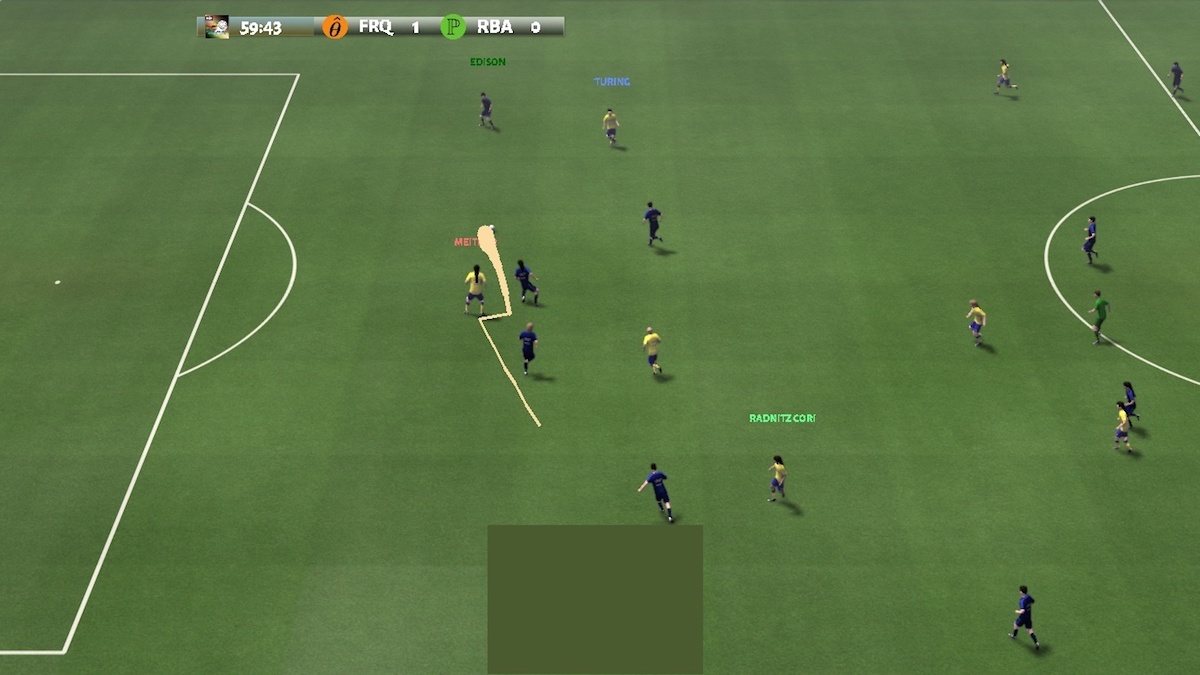}&
\includegraphics[width=0.28\textwidth, trim={0 5.5cm 7cm 0},clip]{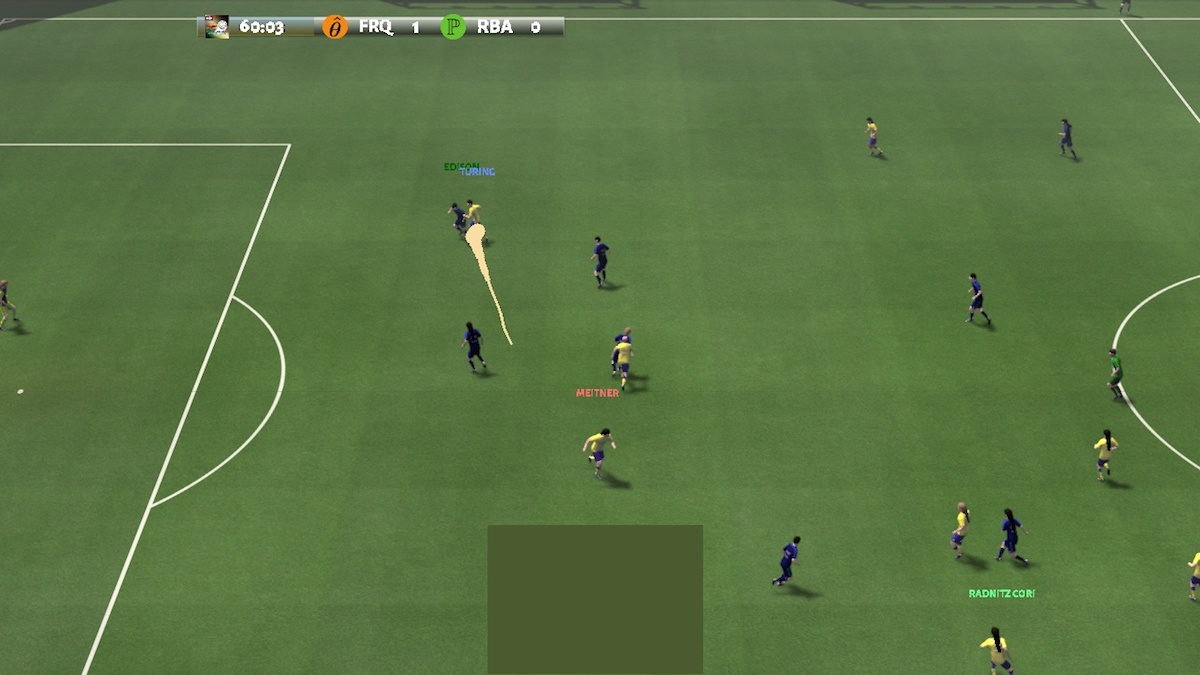}&
\includegraphics[width=0.28\textwidth, trim={0 5.5cm 7cm 0},clip]{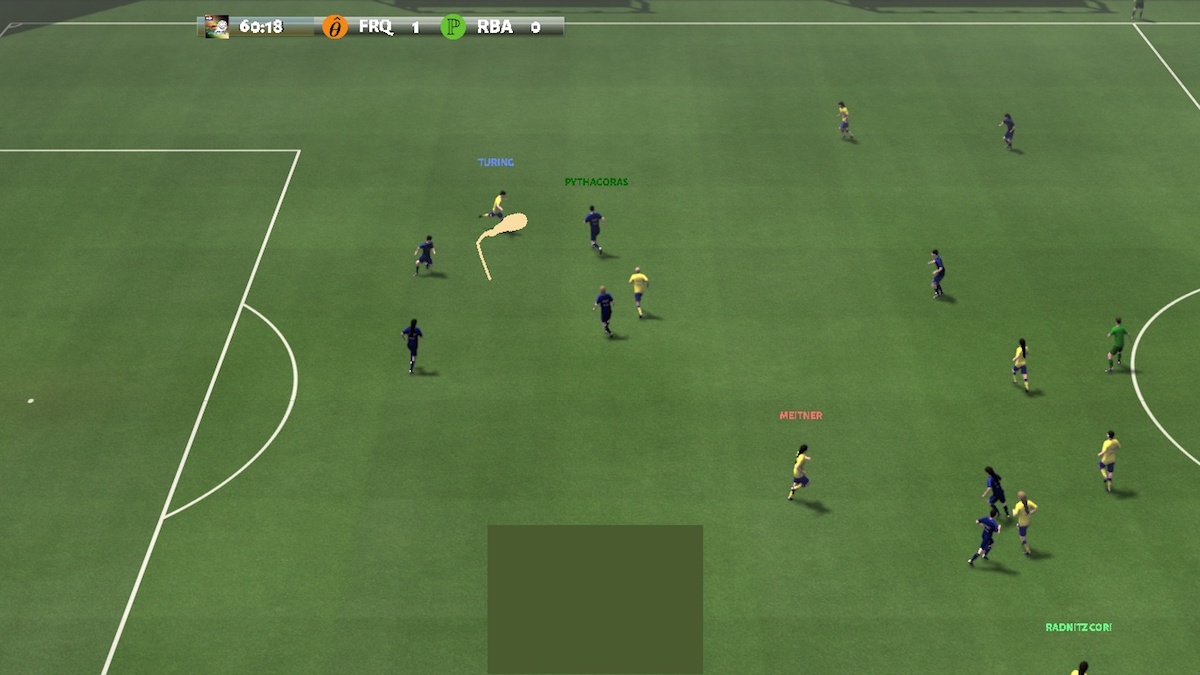}\\
\scriptsize(j) & \scriptsize(k) & \scriptsize(l)\\
\end{tabular}
\vspace{-0.1cm}
\caption{Multi-agent qualitative results for {\em 11 \vs 11} $(N=3)$. (a-f) Fast break. (g-l) Steal and Pass. We visualize the trajectory of the ball in yellow. %
The thickness of the line indicates the time direction, \ie, the thicker the more recent in time. Our policy controlled the team in {\bf \color{gfyy} yellow jerseys}. The controlled players are visualized using a light  {\bf \color{gfrr}  red}, {\bf \color{gfgg} green}, and {\bf \color{gfbb} blue} name tag on top of  each player. 
}
\label{fig:qual_result_multi_football}
\end{figure*}

\subsection{GFootball}
\vspace{-0.1cm}
{\noindent \bf Environment.}
We consider the following tasks in the GFootball environment with up to $N=5$ controlled agents:%

{\em 11 $\vs$ 11:}
Two teams, each of 11 players, play a full 90 minutes game and the episode length is $3000$. We control $N=3$ or $N=5$  agents on the same team aiming to win the football game.

{\em Single goal \vs lazy:} In this task, the opponents cannot move but they can intercept the ball if it is close by. An episode terminates when the agents score or the maximum episode length of $300$ is reached. We control $N=3$ agents. 

{\em 3 \vs 1 with keeper:} Three of our controlled players ($N=3$) try to score from the edge of the box. One player is at the center and the other two players are on the sides. The center player possesses the ball and faces a defender. 

{\em Run, pass and shoot with keeper:} Two of our controlled players ($N=2$) try to score from the edge of the box. One is with the ball and unmarked. The other is at the center, next to a defender, and facing the opponent goal keeper.  

{\em Pass and shoot with keeper:}  Two of our controlled players  ($N=2$) try to score from the edge of the box. One is with the ball and next to a defender. The other one is at the center and facing the opponent goalkeeper.

{\noindent \bf Baselines and metrics.}  We compare to  Kurach \etal~\cite{gfootball} who do not use an intermediate representation. Their policy and value networks are modeled using a CNN taking pixel inputs, \ie, None+CNN. To evaluate our approach, we run 20 evaluation episodes every 100 training episodes. We report the {\em absolute metric}~\cite{Colas18, Henderson17} of the score difference, which is the best policy's average score difference over $20$ evaluation episodes. A positive score difference means the controlled team beats the opponent. Note that in  {\em single goal \vs lazy}, the largest possible score difference is one since the episode terminates when the controlled team scores.

{\noindent \bf Implementation details.}
Following Kurach \etal~\cite{gfootball}, we train both baseline and our approach with parallel PPO~\cite{ppo} using 24 parallel processes. For {\em 11 \vs 11} and {\em single goal \vs lazy} we train the models for $20$M and $5$M environment steps.

In order to keep the RL training time low, detection has to be fast.  We use YOLOv3(+tiny)~\cite{redmon2018yolov3} as the detection framework. The major challenge is detection of small and fast-moving objects like the ball. %
To address this challenge, we use the multi-scale
scheme illustrated in~\figref{fig:det_pipeline}, which finds the ball and the players simultaneously.
Specifically, we detect the players and the ball at two different resolutions as the ball and player sizes differ. To detect the players, we downsample the input image from $720 \times 1280$ to $512\times 512$. 
For detecting the ball, we observe that ball locations are not uniformly distributed as shown in~\figref{fig:ball_heatmap}, where we visualize the probability of the ball at each pixel location in the camera view. 
Leveraging this observation, we perform detection on 
three cropped regions, shown in~\figref{fig:ball_heatmap}, which have a high probability of containing the ball. This permits to avoid a computationally expensive  sliding window method.  We perform post-processing on the results, which includes standard thresholding, non-maximum suppression for player detections, and using the maximum prediction of the ball, as we know a-priori that there is at most one ball in the game. From the detected objects, we then perform tracking to align the objects across frames as visualized in~\figref{fig:tracking_stitch_field}. 

This method detects ball and players at a frame rate of about 50 FPS while having a small memory footprint of around 960 MB. This efficiency permits easy integration of object detection into RL. To train this detector, we collected images with ground-truth bounding boxes from the game engine by running a random policy for 4 episodes (12,000 images). 
Note, this trained detector can be used across different GFootball tasks with differing agent numbers.

{\noindent \bf Visual multi-agent results.}
We report quantitative results in~\tabref{tab:quan_results_football}, where None+CNN is the baseline~\cite{gfootball}, which uses image observations and a CNN. Tracklets+GCN is our approach.  As shown in~\tabref{tab:quan_results_football}, our approach has a $+2.42$ higher score difference than the baseline on the {\em 11 \vs 11} 
task when controlling five visual agents. The training curve of each method is shown in~\figref{fig:quan_results_football}, where we %
averaged over three runs with different random seeds. %
We observe that `semantic tracklets' are more data efficient. {Similarly, for {\em single goal \vs lazy}, {\em 3 \vs 1 with keeper}, {\em Run, pass and shoot with keeper}, and {\em Pass and shoot with keeper}, the approach consistently achieves a higher score difference than the baseline.}

To better understand the learned policies, in~\figref{fig:qual_result_multi_football}, we visualize the learned coordination when controlling three agents. In \figref{fig:qual_result_multi_football} (a-f), our controlled agents pass the ball to players in the front of the court to complete a fast break. In \figref{fig:qual_result_multi_football} (g-l), the controlled agents work together to finish a steal and pass. More specifically, in (g,h) the opponent  (dark jersey) attempts to pass and in (i-l), the controlled player steals the ball and passes the ball to another controlled player. These results demonstrate that our approach learns intricate control of the agents. Please see the supplementary material for videos.

\begin{table}[t]
\centering
\setlength{\tabcolsep}{4.3pt}
\begin{tabularx}{0.9\linewidth}{cccccccc}
\specialrule{.15em}{.05em}{.05em}
\multirow{ 2}{*}{Method}&\multicolumn{7}{c}{Goal difference}\\
& -2.0 & -1.6 & -1.2 & -0.8 & -0.4 &  0.0 & 0.1  \\
\hline
None+CNN~\cite{gfootball} & --& -- & -- & -- & 100 & $>$ 100 & $>$ 100\\
Tracklets &  \textbf{0.2} & \textbf{0.4} & \textbf{1.8} & \textbf{4.0} &\textbf{7.5} &\textbf{15.7} &\textbf{16.0}\\
\specialrule{.15em}{.05em}{.05em}
\end{tabularx}
\vspace{-0.2cm}
\caption{Number of frames, in millions, required to achieve target goal difference for {\em 11 \vs 11} single agent setting. %
}
\label{tab:quan_single_agent_result_football}
\vspace{-0.9cm}
\end{table}
{\noindent \bf Visual single-agent results.} 
We also compare our results with those reported by Kurach \etal~\citet{gfootball} %
in the single agent setting.  As shown in~\tabref{tab:quan_single_agent_result_football}, to achieve a  goal difference of $-0.4$ our approach only needs 7.5M environment steps, whereas None+CNN from Kurach \etal~\cite{gfootball} needs 100M environment steps. Additionally, we achieve a non-negative goal difference within 20M steps. In contrast, the baseline can hardly learn a meaningful policy within 20M environment steps. This demonstrates the data efficiency of  semantic tracklets with GCN: we significantly reduce the number of environment steps to achieve the same reward.

\section{Conclusion}
\label{sec:conc}
We study semantic tracklets, an object-centric  representation for  VMARL. Semantic tracklets permit use of graph-nets and enable efficient learning of policies from visual inputs on  VMPE and GFootball. Notably, for the first time, effective policies are learned for five players in the challenging GFootball setting. Compared to prior work, \eg, not using an intermediate representation or using  segmentation, 
semantic tracklets are a compelling way to improve  VMARL  data efficiency and scalability.

{\small
\bibliographystyle{IEEEtran}
\begingroup
\setlength{\bibsep}{0pt}
\bibliography{example}
\endgroup
}
\clearpage

\end{document}